\documentclass[10pt,twocolumn,letterpaper]{article}

\usepackage[pagenumbers]{cvpr} 

\usepackage{graphicx}
\usepackage{amsmath}
\usepackage{verbatim}
\usepackage{amssymb}
\usepackage{booktabs}
\usepackage{makecell}
\usepackage{multirow}
\usepackage{footnote}
\usepackage[misc]{ifsym}
\usepackage[hang,flushmargin]{footmisc}
\usepackage{graphicx}
\usepackage{amsmath}
\usepackage{algorithm}
\usepackage{algorithmic}
\usepackage{verbatim}
\usepackage{amssymb}
\usepackage{booktabs}
\usepackage{makecell}
\usepackage{multirow}
\usepackage{listings}
\usepackage{bbding}

\usepackage{algorithm}
\usepackage{listings}
\usepackage{dblfloatfix}
\usepackage{etoolbox}
\makeatletter
\AfterEndEnvironment{algorithm}{\let\@algcomment\relax}
\AtEndEnvironment{algorithm}{\kern2pt\hrule\relax\vskip3pt\@algcomment}
\let\@algcomment\relax
\newcommand\algcomment[1]{\def\@algcomment{\footnotesize#1}}
\renewcommand\fs@ruled{\def\@fs@cfont{\bfseries}\let\@fs@capt\floatc@ruled
 \def\@fs@pre{\hrule height.8pt depth0pt \kern2pt}%
 \def\@fs@post{}%
 \def\@fs@mid{\kern2pt\hrule\kern2pt}%
 \let\@fs@iftopcapt\iftrue}
\makeatother

\makeatletter
\newif\if@restonecol
\usepackage[pagebackref,breaklinks,colorlinks]{hyperref}
\usepackage{color, soul}
\setulcolor{blue}
\usepackage{algorithm}
\usepackage{algorithmic}
\usepackage{verbatim}
\usepackage{amsmath} 
\usepackage{float}

\usepackage[pagebackref,breaklinks,colorlinks]{hyperref}

\usepackage[capitalize]{cleveref}
\crefname{section}{Sec.}{Secs.}
\Crefname{section}{Section}{Sections}
\Crefname{table}{Table}{Tables}
\crefname{table}{Tab.}{Tabs.}

\newcommand{\hiddensection}[1]{
    \refstepcounter{section}
    \section*{\arabic{section}.\hspace{1ex}{#1}}
}
\newcommand{\hiddensubsection}[1]{
    \refstepcounter{subsection}
    \subsection*{\arabic{section}.\arabic{subsection}.
    \hspace{1ex}{#1}}
}

\begin{document}

\title{Open-vocabulary Object Segmentation with Diffusion Models\vspace{-0.2cm}}

\author{Ziyi Li$^{1*}$, Qinye Zhou$^{1*}$, Xiaoyun Zhang$^{1}$, Ya Zhang$^{1,2}$, Yanfeng Wang$^{1,2}$, and Weidi Xie$^{1,2}$\\[3pt]
$^{1}$Coop. Medianet Innovation Center, Shanghai Jiao Tong University, China\\[2pt]
$^{2}$Shanghai AI Laboratory, China\\[1pt]
\url{https://lipurple.github.io/Grounded_Diffusion/}
}

\newcommand\blfootnote[1]{%
\begingroup 
\renewcommand\thefootnote{}\footnote{#1}%
\addtocounter{footnote}{-1}%
\endgroup 
}

\twocolumn[{%
\renewcommand\twocolumn[1][]{#1}%
\maketitle


\vspace{-0.8cm}
\begin{center}
    \centering
    \includegraphics[width=.99\linewidth, trim=0 18 0 0]{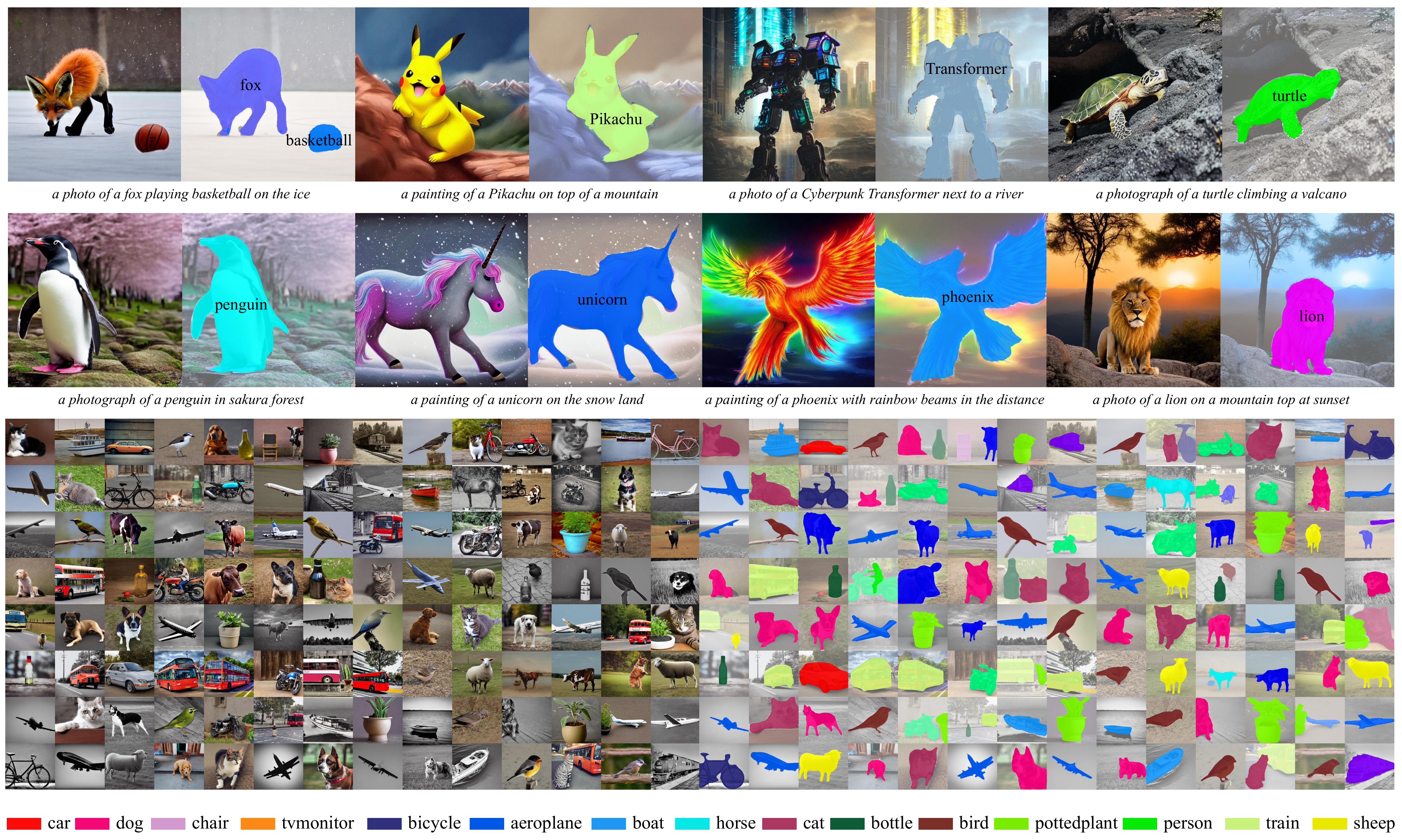}
    \captionof{figure}{\textbf{Predictions from our guided text-to-image diffusion model}. 
    The model is able to simultaneously generate images and segmentation masks for the corresponding visual objects described in the text prompt, for example, \underline{Pikachu}, \underline{Unicorn},  \underline{Phoenix}, {\em etc}.}
\label{fig:teaser}
\end{center}%
}]

\blfootnote{* Both the authors have contributed equally to this project. }

\vspace{-0.3cm}
\begin{abstract}
\vspace{-0.2cm}

The goal of this paper is to extract the visual-language correspondence from a pre-trained text-to-image diffusion model, in the form of segmentation map, 
{\em i.e.}, simultaneously generating images and segmentation masks for the corresponding visual entities described in the text prompt.  
We make the following contributions: 
(i) we pair the existing Stable Diffusion model with a novel grounding module, that can be trained to align the visual and textual embedding space of the diffusion model with only a small number of object categories;
(ii) we establish an automatic pipeline for constructing a dataset, that consists of \{image, segmentation mask, text prompt\} triplets, to train the proposed grounding module; 
(iii) we evaluate the performance of open-vocabulary grounding on images generated from the text-to-image diffusion model and show that the module can well segment the objects of categories beyond seen ones at training time, as shown in Fig.~\ref{fig:teaser};
(iv) we adopt the augmented diffusion model to build a synthetic semantic segmentation dataset, and show that, training a standard segmentation model on such dataset demonstrates competitive performance on the zero-shot segmentation~(ZS3) benchmark, which opens up new opportunities for adopting the powerful diffusion model for discriminative tasks.
\end{abstract}


\hiddensection{Introduction}

In the recent literature, text-to-image generative models have gained increasing attention from the research community and wide public, one of the main advantages of such models is the strong correspondence between visual pixels and language, 
learned from large corpus of image-caption pairs, such correspondence, for instance, enables to generate photorealistic images from the free-form text prompt~\cite{rombach2022high,saharia2022photorealistic,yu2022scaling,ramesh2022hierarchical}. In this paper, we aim to explicitly extract such visual-language correspondence from the generative model in the form of segmentation map, 
{\em i.e.}, simultaneously generating photorealistic images, 
and infer segmentation masks for corresponding visual objects described in the text prompts. The benefit of extracting such visual-language correspondence from generative model is significant, as it enables to synthesize infinite number of images with pixel-wise segmentation for categories within the vocabulary of generative model, 
serving as a free source for augmenting the ability of discriminative segmentation or detection models, {\em i.e.}, be able to process more categories.

To achieve such goal, we propose to pair the existing Stable Diffusion~\cite{rombach2022high} with a novel grounding module, that can segment the corresponding visual objects described in the text prompt, from the generated image. This is achieved by explicitly aligning the text embedding space of desired entity and the visual features of synthetic images,  {\em i.e.}, intermediate layers of diffusion model. Once trained, objects of interest can be segmented by their category names, for both seen and unseen objects at training time, resembling an open-vocabulary object segmentation for generative model.

To properly train the proposed architecture, 
we establish a pipeline for automatically constructing a dataset with \{synthetic image, segmentation mask, text prompt\} triplets, in particular, we adopt an off-the-shelf object detector, and do inference on images generated from the existing Stable Diffusion model, advocating no extra manual annotation.
Theoretically, such a pipeline enables to generate infinite data samples for each category within the vocabulary of existing object detector, for example, we adopt the Mask R-CNN~\cite{liu2021swin} pre-trained on COCO with 80 categories. 
We show that the grounding module trained on a pre-defined set of object categories, can segment images from Stable Diffusion well beyond the vocabulary of any off-the-shelf detector, as shown in Fig.~\ref{fig:teaser}, for example, Pikachu, unicorn, phoenix, {\em etc},
effectively resembling a form of visual instruction tuning, for establishing visual-language correspondence from generative model.

To quantitatively validate the effectiveness of our proposed grounding module, we initiate two protocols for evaluation: {\em first}, we compare the segmentation results with a strong off-the-shelf object detector on synthetic images; {\em second}, we construct a synthesized semantic segmentation dataset with Stable Diffusion and our grounding module, then train a segmentation model on it. While evaluating zero-shot segmentation~(ZS3) on COCO and PASCAL VOC, we outperform prior state-of-the-art models on unseen categories and show competitive performance on seen categories, reflecting the effectiveness of our constructed datasets.
Even more crucially, we demonstrate an appealing application for training discriminative models with synthetic data from generative model, for example, to expand the vocabulary beyond any existing detector can do.

\hiddensection{Preliminary on Diffusion Model}
\label{sec:background}
Diffusion models refer to a series of probabilistic generative models, 
that are trained to learn a data distribution by gradually denoising the randomly sampled Gaussian noises. 
Theoretically, the procedure refers to learning the reverse process of a fixed Markov Chain of length $T$. 
As for text-to-image synthesis, given a dataset of image-caption pairs, 
{\em i.e.}, $\mathcal{D}_{\text{train}} = \{(\mathcal{I}_1, y_1), \dots, (\mathcal{I}_N, y_N)\}$, 
the models can be interpreted as an equally weighted sequence of conditional denoising neural network that iteratively predicts a denoised variant of the input conditioned on the text prompt, 
namely $\epsilon_{\theta}(\mathcal{I}_i^{t},t,y_i)$, 
where $\mathcal{I}_i^t$ denotes a noisy version of the input image,
and $t = 1,\dots,T$ refers to the timestep, $i \in \{1,\dots,N\}$.
For simplicity, we only describe the training and inference procedure for a single image, 
thus ignoring the subscript in the following sections.

In particular, this paper builds on a variant of diffusion model, namely, Stable Diffusion~\cite{rombach2022high}, which conducts the diffusion process in latent space. We will briefly describe its architecture and training procedure in the following. 

\vspace{5pt}
\noindent \textbf{Architecture.}
Stable Diffusion consists of three components: 
a text encoder for producing text embeddings; 
a pre-trained variational auto-encoder that encodes and decodes latent vectors for images; and a time-conditional U-Net~($\epsilon_{\theta}(\cdot)$) for gradually denoising the latent vectors, 
with the progressive convolutional operation that downsamples and upsamples the visual feature maps with skip connections. 
The visual-language interaction occurs in the U-Net via cross-attention layers, specifically, a text encoder is used to project the text prompt $y$ to textual embeddings, that then are mapped into \texttt{Key} and \texttt{Value}, 
and the spatial feature map of the noisy image is linearly projected into \texttt{Query}, iteratively attending the conditioned text prompt for update.

\begin{figure*}[ht]
    \centering
    \includegraphics[width=\linewidth]{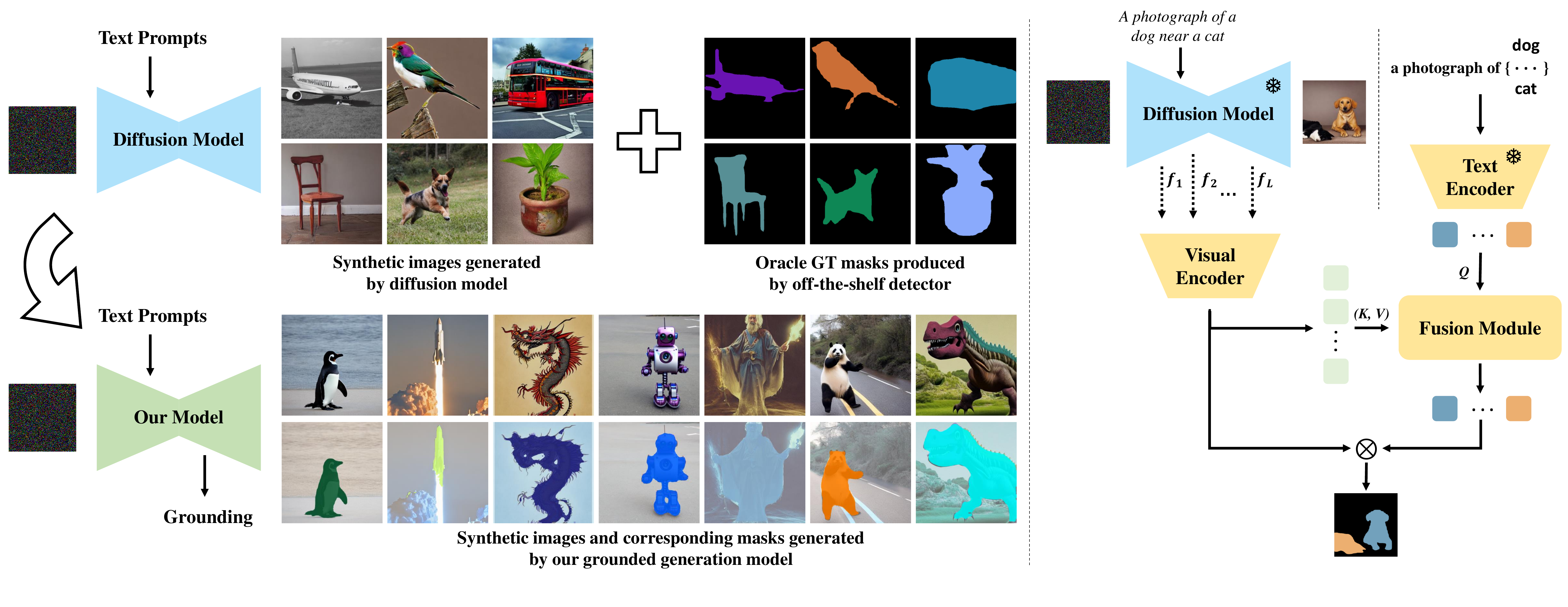}
    \vspace{-0.7cm}
    \caption{\textbf{Overview.} 
    The \textbf{left} figure shows the knowledge induction procedure, where we first construct a dataset with synthetic images from diffusion model and generate corresponding oracle groundtruth masks by an off-the-shelf object detector, which is used to train the open-vocabulary grounding module. 
    The \textbf{right} figure shows the architectural detail of our grounding module, that takes the text embeddings of corresponding entities and the visual features extracted from diffusion model as input, and outputs the corresponding segmentation masks. During training, both the diffusion model and text encoder are kept \textit{frozen}.
    }
    \label{fig:framework}
    \vspace{-0.3cm}
\end{figure*}

\vspace{5pt}
\noindent \textbf{Training and Inference.}
The training procedure of Stable Diffusion can be described as follows: 
given a training pair $(\mathcal{I}, y)$, the input image $\mathcal{I}$ is first mapped to a latent vector $z$ and get a variably-noised vector $z^t:= \alpha^t z + \sigma^t \epsilon $, 
where $\epsilon \sim \mathcal{N}(0,1)$ is a noise term and $\alpha^t, \sigma^t$ are terms that control the noise schedule and sample quality. 
At training time, the time-conditional U-Net is optimised to predict the noise $\epsilon$ and recover the initial $z$, conditioned on the text prompt $y$, the model is trained with a squared error loss on the predicted noise term as follows:
\begin{align}
    \mathcal{L}_{\text{diffusion}} = \mathbb{E}_{z, \epsilon \sim \mathcal{N}(0,1),t,y} \Big[||\epsilon - \epsilon_{\theta}(z^{t},t,y)||_{2}^{2} \Big]
\end{align}
where $t$ is uniformly sampled from $\{1,\dots,T\}$. 


At inference time, Stable Diffusion is sampled by iteratively denoising $z^T \sim \mathcal{N}(0,I)$ conditioned on the text prompt $y$. Specifically, at each denoising step $t = 1,\dots, T$, $z^{t-1}$ is obtained from both $z^t$ and the predicted noise term of U-Net whose input is $z^t$ and text prompt $y$. 
After the final denoising step, $z^0$ will be mapped back to yield the generated image $\mathcal{I}$.

\hiddensection{Problem Formulation}

In this paper, we aim to augment an existing text-to-image diffusion model with the ability of open-vocabulary segmentation, by exploiting the visual-language correspondence, {\em i.e.}, simultaneously generating images, 
and the segmentation masks of corresponding objects described in the text prompt, as shown in Fig.~\ref{fig:framework}~(left):
\begin{align}
\{\mathcal{I}, m\} = \Phi_{\text{diffusion}^{+}}(\epsilon, y)
\end{align}
where $\Phi_{\text{diffusion}^{+}}(\cdot)$ refers to a pre-trained text-to-image diffusion model with a grounding module appended, 
it takes the sampled noise~($\epsilon \sim \mathcal{N}(0,I)$) and language description $y$ as input, and generates an image~($\mathcal{I} \in \mathbb{R}^{H \times W \times 3}$) 
with corresponding segmentation masks~($m \in \{0,1\}^{H \times W \times \mathcal{O}}$) for a total of $\mathcal{O}$ objects of interest. \textbf{Note that}, we expect the model to be {\em open-vocabulary}, 
that means, it should be able to output the corresponding segmentation mask for any objects that can be generated by diffusion model, 
without limitation of the semantic categories.

\hiddensection{Open-vocabulary Grounding}
\label{sec:KI}
Assuming there exists a training set of $N$ triplets, 
{\em i.e.}, $\mathcal{D}_{\text{train}} = \{(\mathcal{F}_1, m_1^{\text{gt}}, y_1), \dots, (\mathcal{F}_N, m_N^{\text{gt}}, y_N)\}$, the predicted segmentation mask for all objects, {\em i.e.},  $m_i \in \mathbb{R}^{H \times W \times \mathcal{O}_i}$ can be obtained by:
\begin{align}
   m_i = \Phi_{\text{fuse}}(\Phi_{\text{v-enc}}(f_i^1, \dots, f_i^n),  \Phi_{\text{t-enc}}(g(y_i)))  
\end{align}
where $y_i$ denotes the text prompt for image generation,
$\mathcal{F}_i = \{f_i^1, \dots, f_i^n\}$ refers to the intermediate representation from Stable Diffusion at $t = 5$~(this has been experimentally validated {in supplementary material}).
Our proposed \textbf{grounding module} consists of three functions, namely,
$\Phi_{\text{v-enc}}(\cdot)$, $\Phi_{\text{t-enc}}(\cdot)$ and $\Phi_{\text{fuse}}(\cdot)$,
denoting visual encoder, text encoder, and fusion module, respectively.
$g(\cdot)$ denotes a group of templates that decorate each of the visual categories in the text prompt, 
{\em e.g.}, `a photograph of a [category name]'.
At training time, there are totally $\mathcal{O}_i$ object categories in the text prompt, and they fall into a pre-defined set of vocabularies $\mathcal{C}_{\text{train}}$;
while at inference time, we expect the model to be highly generalizable, that should equally be capable of segmenting objects from unseen categories, {\em i.e.}, $\mathcal{C}_{\text{test}} \supset \mathcal{C}_{\text{train}}$.

In the following sections, we start by introducing the procedure for automatically constructing a training set in Sec.~\ref{sec:dataset}, followed by the architecture design for open-vocabulary grounding in Sec.~\ref{sec:architecture}, 
lastly, we detail the visual instruction tuning procedure, 
that trains the model with only a handful of image-segmentation pairs as visual demonstrations in Sec.~\ref{sec:training}.



\hiddensubsection{Dataset Construction}
\label{sec:dataset}
Here, we introduce a novel idea to automatically construct \{visual feature, segmentation, text prompt\} triplets, that are used to train our proposed grounding module.
In practise, we first prepare a vocabulary with common object categories, {\em e.g.}, the classes in PASCAL VOC can form a category set as $\mathcal{C}_{\text{pascal}} = \{\text{`dog'}, \text{`cat'}, \dots\}$, $|\mathcal{C}_{\text{pascal}}| = 20$, 
then we randomly select a number of classes to construct text prompts for image generation~({\em e.g.}, `a photograph of a dog and cat'). 
Repeating the above operation, we can theoretically generate an infinite number of image~(intermediate visual representation, {\em i.e.}, $\mathcal{F}$) and text prompt pairs. To acquire the segmentation masks, we take an off-the-shelf object detector, 
{\em e.g.}, pre-trained Mask R-CNN~\cite{liu2021swin}, and run the inference procedure on the generated images:
\begin{align}
m_i^{\text{gt}}=\Phi_{\text{detector}^{}}(\mathcal{I}_i), \text{\hspace{2pt} where }
\mathcal{I}_i = \Phi_{\text{diffusion}}(\epsilon, y_i), 
\end{align}
where $m_i^{\text{gt}} \in \{0,1\}^{H \times W \times \mathcal{O}_i}$ refers to the predicted masks for a total of $\mathcal{O}_i$ objects in the generated image $\mathcal{I}_i$, conditioning on the text prompt $y_i$.

To evaluate the effectiveness of our proposed induction procedure for open-vocabulary grounding, 
we divide the vocabulary into seen categories ($\mathcal{C}_{\text{seen}}$) and unseen categories ($\mathcal{C}_{\text{unseen}}$),
the training set~($\mathcal{D}_{\text{train}}$) only has images of seen categories, 
and the test set~($\mathcal{D}_{\text{test}}$) has both seen and unseen categories. 
\textbf{Note that}, in addition to using the off-the-shelf detector to generate oracle masks and construct dataset, we also explore to utilize \textbf{real public dataset}, {\em e.g.} PASCAL VOC, to train our grounding module via DDIM inverse process. We show the details in Sec.~\ref{sec:DDIMInverse}.

\hiddensubsection{Architecture}
\label{sec:architecture}

Given one specific training triplet, we now detail three  components in the proposed grounding module: visual encoder, text encoder, and fusion module.

\vspace{3pt}
\noindent \textbf{Visual Encoder.}
Given the visual representation,
{\em i.e.}, latent features from the Stable Diffusion,
we concatenate features with the same spatial resolution~(from U-Net encoding and decoding path) to obtain $\{f_i^1, \dots,f_i^n\}$, where $f_i^k \in \mathbb{R}^{\frac{h}{2^k}\times \frac{w}{2^k}\times d_{k}}$, 
$k \in \{0,\dots,n\}$ denotes layer indices of U-Net,
$d_{k}$ refers to the feature dimension.

\begin{figure}[t]
    \centering
    \includegraphics[width=\linewidth]{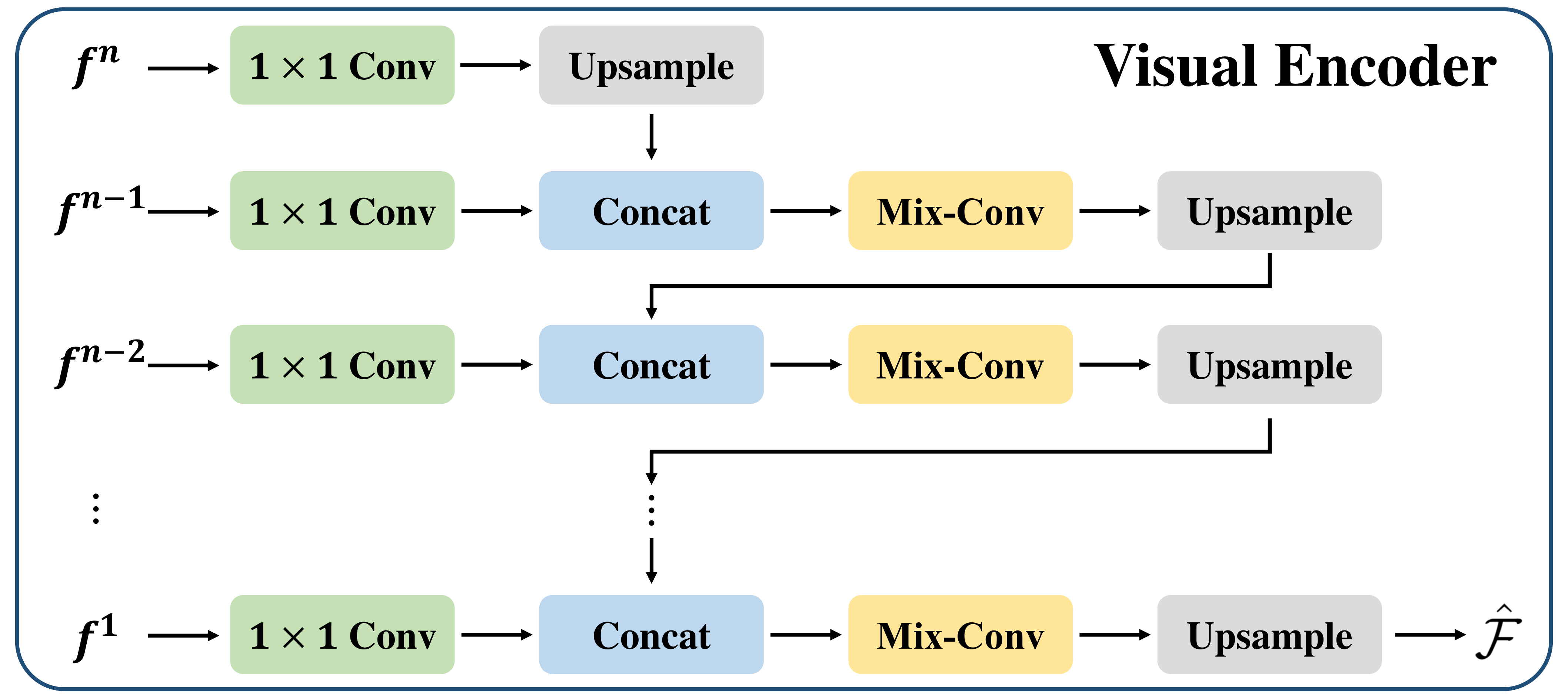}
    \vspace{-0.5cm}
    \caption{\textbf{The architecture of visual encoder.} The features extracted from U-Net are first grouped according to their resolution, then gradually upsampled and fused into the final visual feature.  }
    \label{fig:visual encoder}
    \vspace{-0.3cm}
\end{figure}

Next, we input $\{f_i^1,\dots ,f_i^n\}$ to  visual encoder for generating the fused visual feature $\hat{\mathcal{F}}_i = \Phi_{\text{v-enc}}(\{f_i^1, \dots, f_i^n\})$.
As shown in Fig~\ref{fig:visual encoder}, visual encoder consists of 4 types of blocks: (1) \texttt{$1\times 1$ Conv} for reducing feature dimensionality, 
(2) \texttt{Upsample} for upsampling the feature to a higher spatial resolution,
(3) \texttt{Concat} for concatenating features, and 
(4) \texttt{Mix-conv} for blending features from different spatial resolutions, that includes two \texttt{$3\times 3$ Conv} operations with residual connection and a \texttt{conditional batchnorm} operation, similar to \cite{li2022bigdatasetgan}.

\vspace{5pt}
\noindent \textbf{Text Encoder.}
We adopt the language model from pre-trained Stable Diffusion,
specifically, given the text prompt $y_i$, the text encoder output the corresponding embeddings for all visual objects: $\mathcal{E}_{\text{obj}_i}=\Phi_{\text{t-enc}}(g(y_i))$.

\vspace{5pt}
\noindent \textbf{Fusion Module.}
The fusion module computes interaction between visual and text embeddings, 
then outputs segmentation masks for all visual objects.
In specific, we use a standard transformer decoder with three layers, 
the text embeddings are treated as \texttt{Query}, 
that iteratively attend the visual feature for updating, 
and are further converted into per-segmentation embeddings with a Multi-Layer Perceptron (MLP).
The object segmentation masks can be obtained by dot producting visual features with the per-segmentation embeddings. Formally, the procedure can be denoted as:
\begin{gather}
\mathcal{E}_{\text{seg}_i} = \Phi_{\textsc{Transformer-D}}(W^{Q} {\cdot} \mathcal{E}_{\text{obj}_i}, \hspace{1pt} W^K {\cdot} \hat{\mathcal{F}}_i, \hspace{1pt} W^V {\cdot} \hat{\mathcal{F}_i}) \\
m_i = \hat{\mathcal{F}}_i \cdot [\Phi_{\textsc{MLP}}(\mathcal{E}_{\text{seg}_i})]^T
\end{gather}
where the transformer decoder generates per-segmentation embedding $\mathcal{E}_{\text{seg}_i} \in \mathbb{R}^{N \times d_e}$ for all visual objects described in the text prompt, $W^Q, W^K, W^V$ refer to the learnable parameters for \texttt{Query}, \texttt{Key} and \texttt{Value} projection.

\begin{table*}[t]
\footnotesize
\begin{center}
\tabcolsep=0.2cm
\begin{tabular}{cc|ccccc|ccccc}
\cmidrule(){1-12}
\multirow{4}{*}{\makecell[c]{Test\\Setting}} &  & \multicolumn{5}{c|}{\textbf{PASCAL-sim}}  & \multicolumn{5}{c}{\textbf{COCO-sim}} \\
\cmidrule( ){2-12}
  & \# Objects & \multicolumn{2}{c|}{One}  & \multicolumn{3}{c|}{Two} & \multicolumn{2}{c|}{One}  & \multicolumn{3}{c}{Two}\\
 \cmidrule( ){2-12}
 & Categories & \makecell[c]{Seen} & \multicolumn{1}{c|}{Unseen}
  & \makecell[c]{Seen} & \makecell[c]{Seen +Unseen} & \multicolumn{1}{c|}{Unseen} & \makecell[c]{Seen} & \multicolumn{1}{c|}{Unseen}  & \makecell[c]{Seen} & \makecell[c]{Seen +Unseen} & \multicolumn{1}{c}{Unseen} \\
\cmidrule( ){1-12}
\multirow{4}{*}{DAAM~\cite{tang2022daam}} & $\mathrm{Split} 1$ & 61.66 &\multicolumn{1}{c|}{75.63} &46.74 & 51.31& 69.94&62.25& \multicolumn{1}{c|}{55.56}&49.68 &52.06 &43.35  \\
 & $\mathrm{Split} 2$ & 65.75&\multicolumn{1}{c|}{59.25} &49.08 & 47.98& 41.50&60.08 & \multicolumn{1}{c|}{65.55}&48.80 &54.66&33.22  \\
 & $\mathrm{Split} 3$ &67.11 &\multicolumn{1}{c|}{53.82} & 48.80 & 48.28 &41.41& 62.81& \multicolumn{1}{c|}{52.48}&50.85 &49.84 & 45.80 \\
\cmidrule( ){2-12}
 & Average &64.84 & \multicolumn{1}{c|}{62.90}& 48.21&49.19 & 50.95& 61.71& \multicolumn{1}{c|}{57.76}& 49.78&52.19 & 40.79 \\
\cmidrule( ){1-12}
\multirow{4}{*}{Ours} & $\mathrm{Split} 1$ & 90.16 &\multicolumn{1}{c|}{83.19} &78.93 & 66.07& 57.93&83.35 & \multicolumn{1}{c|}{76.81}&64.64 &57.15 &47.77  \\
 & $\mathrm{Split} 2$ & 90.08&\multicolumn{1}{c|}{86.19} &78.68 & 67.10& 47.21&82.83 & \multicolumn{1}{c|}{74.93}&63.39 &57.18&42.82  \\
 & $\mathrm{Split} 3$ &90.67 &\multicolumn{1}{c|}{79.86} & 79.68 & 70.42 &62.07& 84.85& \multicolumn{1}{c|}{67.89}&65.70 &54.60 & 42.62 \\
\cmidrule( ){2-12}
 & Average &\textbf{90.30} & \multicolumn{1}{c|}{\textbf{83.08}}& \textbf{79.10}&\textbf{67.86} & \textbf{55.74}& \textbf{83.68}& \multicolumn{1}{c|}{\textbf{73.21}}& \textbf{64.16}&\textbf{56.31} & \textbf{44.40} \\   
\cmidrule( ){1-12}

\end{tabular}
\end{center}
\vspace{-0.6cm}
\caption{\textbf{Quantitative result for Protocol-I evaluation on grounded generation.} 
Our model has been trained on the synthesized training dataset, that consists of images with one or two objects from only seen categories, and test on our synthesized test dataset that consists of images with one or two objects of both seen and unseen categories. Split1, Split2 and Split3 refer to 3 different splits of $\mathcal{C}$ that construct 3 different $(\mathcal{C}_\text{seen},\mathcal{C}_\text{unseen})$ pairs, respectively.
Our model outperforms the DAAM~\cite{tang2022daam}, by a large margin, see text for more detailed discussion.
}
\label{tab:main-ex1}
\vspace{-0.4cm}
\end{table*}

\hiddensubsection{Training}
\label{sec:training}

With the constructed dataset, we can now start supervised training the proposed grounding module:
\begin{small}
\begin{align*}
    \mathcal{L}= -\frac{1}{N}\sum_{i=1}^N[m_i^{\text{gt}}\cdot \log(\sigma(m_i))+(1-m_i^{\text{gt}})\cdot \log(\sigma(1-m_i))]
\end{align*}
\end{small}

\noindent where $m^{gt}_i \in \mathbb{R}^{H \times W \times \mathcal{O}_i}$ refers to the oracle groundtruth segmentation from the off-the-shelf object detector, and $m_i \in \mathbb{R}^{H \times W \times \mathcal{O}_i}$ refers to the predicted segmentation from our grounding module, 
$\sigma(\cdot)$ refers to Sigmoid function.

In practise, while using the off-the-shelf detector to generate segmentation masks, 
the model may sometimes fail to detect the objects mentioned in the text prompt, 
and output incorrect segmentation masks.
Such error comes from two sources,
(i) the diffusion model may fail to generate high-quality images;
(ii) off-the-shelf detector may fail to detect the objects in the synthetic image, due to the domain gap between synthetic and real images. 
Here, we consider two training strategies, \textbf{Normal Training}, where we fully trust the off-the-shelf detector, and use all predicted segmentation masks to train the grounding module;
alternatively, we also try \textbf{Training w.o.~Zero Masks}, 
as we empirically found that the majority of failure cases come from false negatives, that is to say, the detector failed to detect the objects and output an all-zero mask, therefore, we can train the grounding modules by ignoring the all-zero masks.

\hiddensection{Experiments}
In this section, we present the evaluation detail for validating the effectiveness of grounding module and its usefulness for training discriminative models, specifically, we consider two protocols: 
in Sec.~\ref{section5.1}, we train the grounding module with the constructed training set, and test the segmentation performance on synthetically generated images from Stable Diffusion, 
then compare with a strong off-the-shelf object detector;
in Sec.~\ref{section:5.2}, we use our augmented diffusion model to construct a synthesized semantic segmentation dataset and train a semantic segmentation model for zero-shot segmentation. 
Lastly, in Sec.~\ref{section:ablation}, 
we conduct a series of ablation studies on the different training strategies and effects on the number of seen categories.

\begin{figure*}[t]
    \centering
    \includegraphics[width=\linewidth]{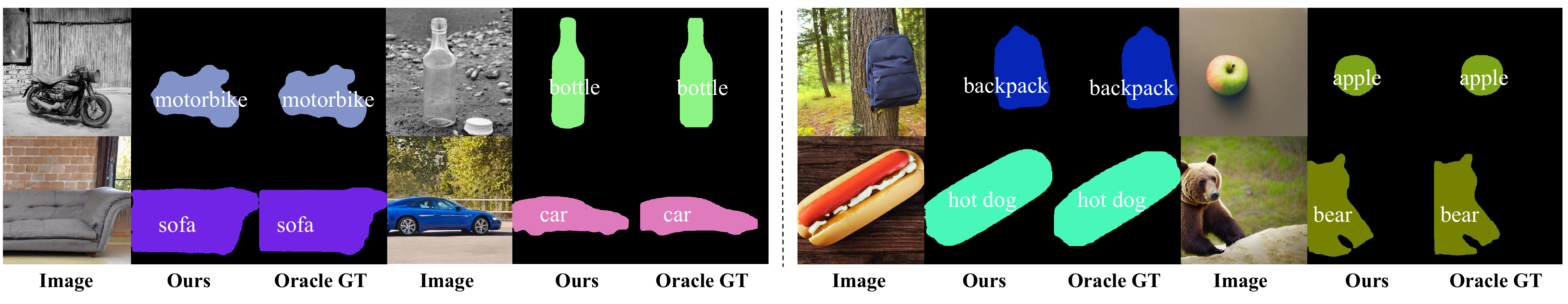}
    \vspace{-0.8cm}
    \caption{\textbf{Segmentation results of PASCAL-sim (left) and COCO-sim (right)}  on seen (motorbike, bottle, backpack, and apple) and unseen (sofa, car, hot dog, and bear) categories. Our grounded generation model achieves comparable segmentation results to the oracle groundtruth generated by the off-the-shelf object detector.  }
    \label{fig:result_pascal}
\end{figure*}


\begin{figure*}[t]
  \centering
  \includegraphics[width=\linewidth]{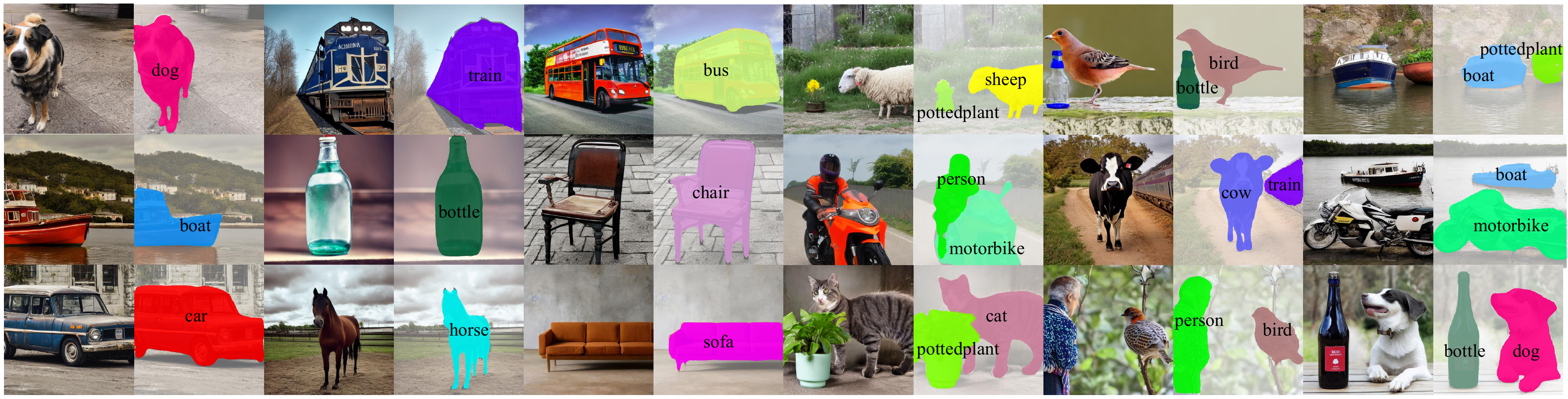}
  \vspace{-0.6cm}
  \caption{\textbf{Our synthesized semantic segmentation dataset} with one category (left) and two categories (right) for Protocol-II training.}
\label{fig:dataset}
\vspace{-0.2cm}
\end{figure*}


\hiddensubsection{Protocol-I: {Open-vocabulary Grounding}}
\label{section5.1}
Here, we train the grounding module with our constructed training set, as described in Sec.~\ref{sec:dataset}. Specifically, the training set only consists of a subset of common~(seen) categories, 
while the test set consists of both seen and unseen categories. 
In the following, we describe the implementation and experimental results in detail, to thoroughly assess the model on open-vocabulary grounding.

Following the dataset construction procedure as introduced in Sec.~\ref{sec:dataset}, 
we make \textbf{PASCAL-sim} and \textbf{COCO-sim}, 
{with the same category split of PASCAL VOC~\cite{everingham2010pascal} and COCO~\cite{lin2014microsoft} as in ~\cite{gu2020context,xian2019semantic,ding2022decoupling}}:
(i) PASCAL-sim contains 15 seen and 5 unseen categories respectively;
(ii) COCO-sim has 65 seen and 15 unseen categories respectively.

\vspace{3pt}
\noindent \textbf{Training Set.}
For PASCAL-sim or COCO-sim, we generate a synthetic training set by randomly sampling images from pre-trained Stable Diffusion.
This exposes our grounding module to a great variety of data ($>$ 40k) at training time, and the model is unlikely to see the same labeled examples twice during training. 
Unlike BigDatesetGAN~\cite{li2022bigdatasetgan}, 
where only a single object is considered, 
we construct the text prompt with one or two objects at a time, 
note that, for training, only the seen categories are sampled. 
Although we can certainly generate images with more than two object categories, 
the quality of the generated images tends to be unstable, limited by the generation ability of Stable Diffusion, thus we only consider synthesized images with less than three object categories. 

\vspace{3pt}
\noindent \textbf{Testing Set.}
For evaluation purpose, 
we generate two synthetic test sets with offline sampling for PASCAL-sim and COCO-sim, respectively.
In total, we have collected about 1k images for PASCAL-sim, 
and about 5k images for COCO-sim, we run the off-the-shelf object detector on these generated images to produce the oracle segmentation. 
For both test sets, the images containing two categories will be divided into three groups: 
both seen, both unseen, one seen and one unseen. 
We leave the detailed statistics of our synthetic dataset in the supplementary material. 
\textbf{Note that}, we have manually checked all the images and the segmentation produced from the off-the-shelf detector, and only keep the high-quality ones, thus the performance evaluation of our grounding module can be a close proxy.

\vspace{3pt}
\noindent \textbf{Evaluation Metrics.}
We use the category-wise mean intersection-over-union (mIoU) as evaluation metric,
defined as averages of IoU over all the categories: 
mIoU $= \frac{1}{C} \sum^{C}_{c=1}\mathrm{IoU}_{c}$, where
$C$ is the number of all target categories, 
and $\mathrm{IoU}_{c}$ is the intersection-over-union for the category with index is $c$. 

\vspace{3pt}
\noindent \textbf{Baseline.}
DAAM~\cite{tang2022daam} is used as a baseline for comparison, where the attention maps 
are directly upsampled and aggregated at each time step to explore the influence area of each word in the input text prompt.

\vspace{3pt}
\noindent \textbf{Implementation Details.}
We use the pre-trained Stable Diffusion~\cite{rombach2022high} and the text encoder of CLIP~\cite{radford2021learning} in our implementation. 
We choose the Mask R-CNN~\cite{liu2021swin} trained on COCO dataset as our object detector for oracle groundtruth segmentation. 
We fuse features from U-Net and upsample them into $512\times 512$ spatial resolution,
the text and visual embeddings are both mapped into 512 dimensions before feeding into the fusion module. 
We train our grounding module with two NVIDIA GeForce RTX 3090 GPUs for 5k iterations with batch size equal to 8, ADAM\cite{kingma2014adam} optimiser with $\beta_{1} = 0.9$, $\beta_{2} = 0.999$. 
The initial learning rate is set to 1e-4 and the weight decay is 1e-4.

\vspace{3pt}
\noindent \textbf{Results.}
As shown in Tab.~\ref{tab:main-ex1}, we provide experimental results for our grounding module, we change the composition of categories three times and compute the results for each split. 
Here, we can make the following observations: 
{\em first}, our model significantly outperforms the unsupervised method DAAM~\cite{tang2022daam} in the mIoU on all test settings. This is because DAAM tends to result in ambiguous segmentations, as the textual embedding for individual visual entity will largely be influenced by {other entities and  the global sentence} at the text encoding stage; 
{\em second}, our grounding module achieves superior performance on both seen and unseen categories, indicating its open-vocabulary nature, {\em i.e.}, the guided diffusion model can synthesize images and their corresponding segmentations for more categories beyond the vocabulary of the off-the-shelf detector.

\begin{table*}[t]
\footnotesize
\begin{center}
\tabcolsep=0.18cm
\begin{tabular}{c|cc|ccc|cc|ccc}
\cmidrule(){1-11}
              & \multicolumn{5}{c|}{ PASCAL VOC}   &    \multicolumn{5}{c}{COCO}     \\
              \cmidrule(){1-11}
\multirow{3}{*}{\makecell[c]{\textbf{Method}}} & \multicolumn{2}{c|}{Training  Set / \# Categories } & \multicolumn{3}{c|}{mIOU} & \multicolumn{2}{c|}{Training  Set / \# Categories }  & \multicolumn{3}{c}{mIOU}   \\ \cmidrule(){2-11}
                         & \multicolumn{1}{c}{Real / 15   }       & 
                         \multicolumn{1}{c|}{\makecell[c]{Synthetic /  15+5 \\(\# Objects) }}    
                         &                       Seen & Unseen & Harmonic   
                         & \multicolumn{1}{c}{Real / 65  }      
                         & \multicolumn{1}{c|}{\makecell[c]{Synthetic / 65+15\\ (\# Objects) } }     
                         &                    Seen & Unseen & Harmonic    \\

                 \cmidrule(){1-11}
ZS3~\cite{bucher2019zero}            &\Checkmark&  \XSolidBrush  & 78.0        & 21.2   & 33.3  &\Checkmark& \XSolidBrush  & -        & -   & -   \\
SPNet~\cite{xian2019semantic}       &\Checkmark&  \XSolidBrush  & 77.8        & 25.8   & 38.8  &\Checkmark&  \XSolidBrush   & 33.0       & 21.9  & 26.3  \\
CaGNet~\cite{gu2020context}         &\Checkmark& \XSolidBrush & 78.6        & 30.3   & 43.7   &\Checkmark& \XSolidBrush & -        & -   & -   \\
Joint~\cite{baek2021exploiting}       & \Checkmark & \XSolidBrush  & 77.7        & 32.5   & 45.9  & \Checkmark & \XSolidBrush   & \textbf{57.9}        & 8.6   & 14.9   \\
STRICT~\cite{pastore2021closer}     &\Checkmark&    \XSolidBrush   & 82.7        & 35.6   & 49.8  &\Checkmark&   \XSolidBrush   & 22.2       & 20.4   & 21.3  \\
SIGN~\cite{cheng2021sign}           &\Checkmark& \XSolidBrush &  \underline{83.5}       & 41.3   & 55.3  &\Checkmark&  \XSolidBrush &  -       & -   & -   \\
ZegFormer\cite{ding2022decoupling} &\Checkmark &\XSolidBrush &  \textbf{86.4}        & \textit{63.6}   & \textit{73.3}  &\Checkmark &\XSolidBrush &  \textit{53.3}    & 34.5   
& \textit{41.9}\\
\cmidrule(){1-11}
Model-A~(Ours)   & \XSolidBrush & \Checkmark (one) &   62.8          &    50.0    &  55.7  &\XSolidBrush & \Checkmark (one) &   28.8          &    32.6    &  30.6     \\
Model-B~(Ours)   & \XSolidBrush& \Checkmark (two) &  65.8           &  60.1      & 62.8  & \XSolidBrush& \Checkmark (two) &  37.3           &  37.0      & 37.1      \\
Model-C~(Ours)   &\XSolidBrush & \quad ~ \Checkmark (mixture) &     69.5        &   63.2    &  66.2  & \XSolidBrush& \quad ~ \Checkmark (mixture) &  38.3           &  \underline{38.1}      & 38.2     \\
Model-D~(Ours)   &\Checkmark& \quad ~ \Checkmark (mixture) &     83.0        &   \underline{71.3}    &  \underline{76.7}  &\Checkmark & \quad ~ \Checkmark   (mixture) &  50.0           &  \textbf{38.2}      & \underline{43.2}     \\
\cmidrule(){1-11}
{\makecell[c]{Model-E~(Ours) \\ (complicated prompt) }}   &\Checkmark& \quad ~ \Checkmark (mixture) &     \textit{83.4}        &   \textbf{74.4}    &  \textbf{78.7}  &\Checkmark & \quad ~ \Checkmark   (mixture) &  \underline{53.4}           &  \textit{37.8}      & \textbf{44.3}     \\
\cmidrule(){1-11}
\end{tabular}
\vspace{-0.2cm}
\caption{
{\textbf{Comparison with previous ZS3 methods on the test sets of PASCAL VOC and COCO.} These ZS3 methods are trained on real training sets. 
The results on PASCAL VOC are from \cite{ding2022decoupling}. 
And we retrained these ZS3 methods on COCO using their official codes, 
however, due to missing implementation details, we failed to reproduce ZS3~\cite{bucher2019zero}, CaGNet~\cite{gu2020context} and SIGN~\cite{cheng2021sign}. }}
\label{tab:ex3_4}
\end{center}
\vspace{-0.5cm}
\end{table*}

\vspace{3pt}
\noindent \textbf{Visualization.}
We demonstrate the visualization results in Fig.~\ref{fig:result_pascal}. 
On both seen and unseen categories, 
our model can successfully ground the objects in terms of segmentation mask. Impressively, as shown in Fig.~\ref{fig:teaser}, 
our grounding module can even segment the objects 
beyond any off-the-shelf detector can do, showing the strong open-vocabulary grounding ability of our model.

\vspace{3pt}
\hiddensubsection{Protocol-II: Open-vocabulary Segmentation}
\label{section:5.2}
In the previous protocol, we have validated the ability for open-vocabulary grounding on synthetically generated images.
Here, we consider to adopt the images and grounding masks for tackling discriminative tasks. 
In particular, we first construct a synthesized image-segmentation dataset with the Stable Diffusion and our grounding module, 
then train a standard semantic segmentation model on such a synthetic dataset, and evaluate it on public image segmentation benchmarks.


\begin{figure*}[t]
    \centering
    \includegraphics[width=\linewidth]{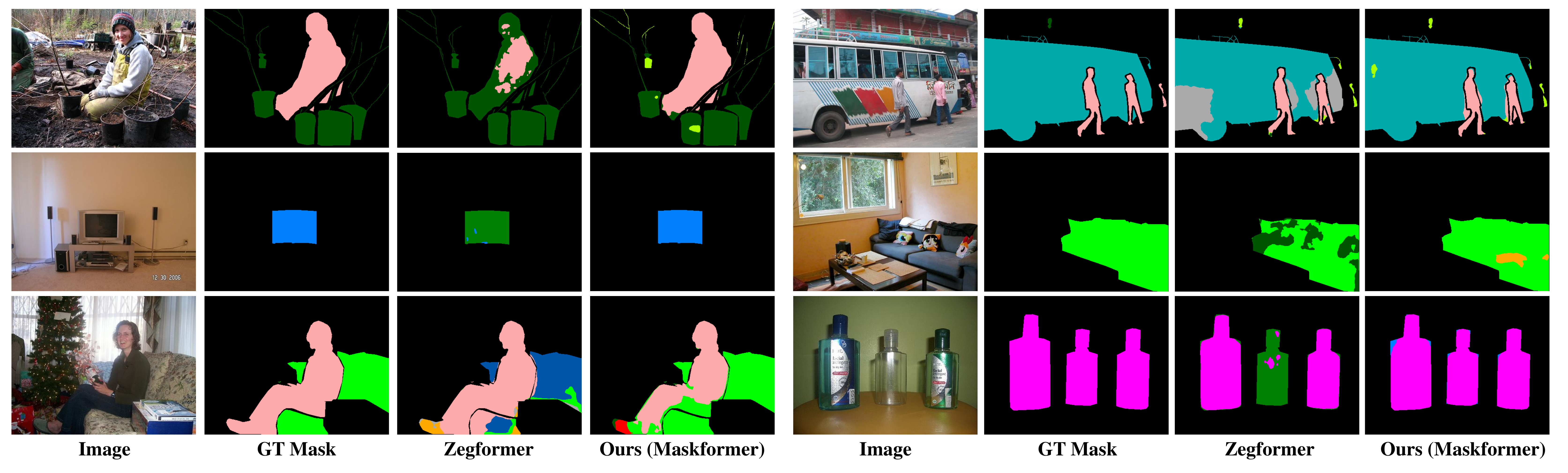}
    \vspace{-0.7cm}
    \caption{
    {\textbf{Qualitative results compared with the state-of-the-art zero-shot semantic segmentation method (Zegformer) on Pascal VOC.} 
    While training on our synthetic dataset and real dataset (only PASCAL VOC seen) together, a standard MaskFormer architecture can achieve better performance than Zegformer on unseen categories, 
    {\em i.e.} pottedplant, sofa and tvmonitor.  
    \textbf{Note that}, the synthetic data for training MaskFormer are generated from our grounding module, which has only been trained on seen categories, with no extra manual annotation involved whatsoever.
    }}
    \label{fig:visual zs3}
    \vspace{-0.4cm}
\end{figure*}

\vspace{2pt}
\noindent \textbf{Synthetic Dataset.}
To train the semantic segmentation model, 
in addition to the provided datasets from public benchmarks, 
we also include two synthetically generated datasets: 
(i) 10k image-segmentation pairs for 20 categories in PASCAL VOC;
(ii) 110k pairs for 80 categories in COCO, as shown in Fig.~\ref{fig:dataset}. All the image-segmentation pairs are generated by the Stable Diffusion and our proposed grounding module,
that has only been trained on the corresponding seen categories~(15 seen categories in PASCAL VOC and 65 seen categories in COCO). 
That is to say, generating these two datasets requires \textbf{no extra manual annotations} whatsoever.

\vspace{2pt}
\noindent \textbf{Training Details.}
To compare with other open-vocabulary methods, 
our semantic segmentation model uses MaskFormer~\cite{cheng2021per} with ResNet101 as its backbone. 
The image resolution for training is 224$\times$224 pix, 
and we train the model on our synthetic dataset for 40k iterations with  batch size equal to 8. 
We use the ADAMW as our optimizer with a learning rate of 1e-4 and the weight decay is 1e-4.

\vspace{2pt}
\noindent \textbf{Comparison on Zero-Shot Segmentation~(ZS3).}
While evaluating on the test sets of \textbf{real} images~(1,449 images for PASCAL VOC and 5000 images for COCO),
we compare with the existing zero-shot semantic segmentation approaches. 
As shown in Tab.~\ref{tab:ex3_4},
while only trained on synthetic dataset,
our model-A,B,C have already outperformed most of ZS3 approaches on unseen categories. Specifically, the model-C trained on the mixture of one and two objects achieves the best performance. 
Additionally, finetuning on the real dataset with images of seen categories can further improve the performance, especially on seen categories~(model-D). We also try to construct the synthetic dataset with more complicated prompt, {\em e.g.}, `A photograph of a \textit{bus} crossing a busy intersection in Seoul's bustling city center', and the model-E gets a slight boost on final performance.
Qualitative results can be seen in Fig.~\ref{fig:visual zs3}, 
our model obtains accurate semantic segmentation on both seen and unseen categories.



\vspace{4pt}
\noindent \textbf{Discussion.}
Overall, we can draw the following conclusions: (i) the grounding module is capable of segmenting unseen categories despite it has never seen any segmentation mask during the knowledge induction procedure, validating the strong generalisation of the grounding module; 
(ii) it is possible to segment more object categories by simply training on synthesized datasets, and the addition of real datasets with only seen categories can narrow the data gap thus resulting in better performance;
(iii) with our proposed idea for extracting the visual-language correspondence from  generative model, it introduces promising applications for applying the powerful diffusion model for discriminative tasks, {\em i.e.}, constructing dataset with generative model, and use it to train discriminative models, {\em e.g.}, expand the vocabulary of an object segmentor or detector.

\begin{table}[t]
\footnotesize
\begin{center}
\tabcolsep=0.07cm
\begin{tabular}{cccccc}
\cmidrule( ){1-6}
 \multirow{2}{*}{Training Type} & \multicolumn{2}{c}{One}  & \multicolumn{3}{c}{Two} \\
 \cmidrule(lr){2-3}\cmidrule(lr){4-6}
 & \makecell[c]{Seen} & \multicolumn{1}{c}{Unseen}
  & \makecell[c]{Seen} & \makecell[c]{Seen +Unseen} & \multicolumn{1}{c}{Unseen} \\
\cmidrule( ){1-6}
Normal Training &89.88 &\multicolumn{1}{c}{71.18} & 77.66& 57.24& 44.22  \\
Training w.o.~Zero Masks & \textbf{90.16} &\multicolumn{1}{c}{\textbf{83.19}} &\textbf{78.93} & \textbf{66.07}& \textbf{57.93}  \\
\cmidrule( ){1-6}
\end{tabular}
\end{center}
\vspace{-0.7cm}
\caption{\textbf{Ablation on training type on the constructed dataset.} Performance is measured by mIoU on \textbf{PASCAL-sim} test set. }
\label{tab:ablation-1}
\vspace{-0.4cm}
\end{table}


\hiddensubsection{Ablation study}
\label{section:ablation}
In this section, we show the effect of different training loss and different numbers of seen categories. Due to the space limitation, we refer the reader for supplementary material, for the study on the different timestep for extracting visual representation, the number of objects in the synthetic datasets, and the effect of different datasets.

\vspace{3pt}
\noindent \textbf{Normal Training v.s.~Training without Zero Masks.}
As shown in Tab.~\ref{tab:ablation-1},
\textbf{Normal Training} results in unsatisfactory performance on unseen categories, we conjecture this is because the errors from detector tend to be false negative, that bias our grounding module to generate all-zero segmentation masks when encountering unseen categories; in contrast, by ignoring all-zero masks at training, \textbf{Training w.o.~Zero Masks} achieves equally good performance on both seen and unseen categories. 

\vspace{3pt}
\noindent \textbf{Effect on the Number of Seen Categories.}
We ablate the number of seen categories to further explore the generalisation ability of our proposed grounding module.
As shown in Tab.~\ref{tab:ablation-3}, 
the grounding module can generalise to unseen categories, 
even with as few as five seen categories;
when introducing more seen categories, 
the performance on unseen ones consistently improves, 
but decreases on seen ones, due to the increasing complexity on seen categories.






\hiddensection{Related Work}

\vspace{2pt}
\noindent \textbf{Image Generation.}
Image generation is one of the most challenging tasks in computer vision due to the high-dimensional nature of images. 
In the past few years, generative adversarial networks~(GAN)~\cite{goodfellow2020generative}, 
variational autoencoders~(VAE)~\cite{kingma2013auto}, 
flow-based models~\cite{kingma2018glow} and autoregressive models~(ARM)~\cite{van2016conditional} have made great progress. 
However, even GANs, the best of these methods, still face training instability and mode collapse issues~\cite{brock2018large}. 
Recently, Diffusion Probabilistic Models (DM) demonstrate state-of-the-art generation quality on highly diverse datasets~\cite{ho2020denoising,nichol2021improved,saharia2022image,ho2022cascaded,saharia2022palette}, outperforming GANs in fidelity~\cite{dhariwal2021diffusion}. These models are often combined with a well-designed text encoder and trained on billions of image-caption pairs for text-to-image generation task, {\em i.e.}, OpenAI’s DALL-E 2~\cite{ramesh2022hierarchical}, 
Google’s Imagen~\cite{saharia2022photorealistic} and Stability AI's Stable Diffusion~\cite{rombach2022high}. 
However, despite being able to generate images with impressive quality using free-form text, 
it remains unclear what extent the visual-language correspondence has been successfully captured,
this paper aims to augment an existing text-to-image diffusion model with the ability to ground objects in its generation procedure.

\vspace{3pt}
\noindent \textbf{Visual Grounding.}
Visual grounding, also known as referring expression comprehension, 
expects to understand the natural language query and then find out the target object of the query in an image. Early visual grounding works are trained in two stages~\cite{hong2019learning,liu2019learning,wang2018learning,wang2019neighbourhood,yang2019dynamic}, 
by first detecting the candidate regions, and then ranking these regions. 
Later, one-stage approaches~\cite{liao2020real,sadhu2019zero,yang2020improving,yang2019fast} attract more attention due to their superior accuracy and efficiency in fusing linguistic context and visual features.
Here, we consider visual grounding in the image generation procedure.

\hiddensection{Conclusion}

\begin{table}[t]
\footnotesize
\begin{center}
\tabcolsep=0.15cm
\begin{tabular}{ccccccc}
\cmidrule( ){1-6} \multirow{3}{*}{\makecell[c]{Train Set \\ \# Seen / Unseen }} & \multicolumn{2}{c}{One}  & \multicolumn{3}{c}{Two} \\
 \cmidrule(lr){2-3}\cmidrule(lr){4-6}
 &  \makecell[c]{Seen} & \multicolumn{1}{c}{Unseen}
  & \makecell[c]{Seen} & \makecell[c]{Seen +Unseen} & \multicolumn{1}{c}{Unseen} \\
\cmidrule( ){1-6}
 5 \quad / \quad 75 & \textbf{94.81} &\multicolumn{1}{c}{72.42} &\textbf{87.19} &49.60 & 39.00 \\
 20 \quad / \quad 60 & 91.91 &\multicolumn{1}{c}{73.33} &71.59 & 56.27&   41.91\\
 35 \quad / \quad 45  & 87.23 &\multicolumn{1}{c}{73.85} &66.91 & 55.99&  43.28\\
 50 \quad / \quad 30 & 84.55 &\multicolumn{1}{c}{73.20} &66.41 & 54.39& 42.71\\
 65 \quad / \quad 15 & 83.85 &\multicolumn{1}{c}{\textbf{76.81}} &64.64 &\textbf{57.15} &  \textbf{47.77}\\
\cmidrule( ){1-6}
\end{tabular}
\end{center}
\vspace{-0.5cm}
\caption{\textbf{Ablation on the number of seen categories on COCO-sim.} The bolded number indicates the best result. Our model can generalise to unseen categories, 
even as few as five seen categories. }
\label{tab:ablation-3}
\vspace{-0.4cm}
\end{table}

In this paper, 
we propose a novel idea for guiding the existing Stable Diffusion towards {grounded generation}, {\em i.e.}, segmenting the visual entities described in the text prompt while generating images. 
Specifically, we introduce a grounding module that explicitly aligns the visual and textual embedding space of the Stable Diffusion and train such module with an automatically constructed dataset,
consisting of \{image, segmentation, text prompts\} triplets.
Experimentally, we show that visual-language correspondence can be established by only training on a limited number of object categories, while getting the ability for open-vocabulary grounding at the image generation procedure. Additionally, we generate a synthetic semantic segmentation dataset using the augmented Stable Diffusion and train a semantic segmentation model.
{The model can transfer to real images and show competitive performance to existing zero-shot semantic segmentation approaches on PASCAL VOC and COCO dataset, opening up new opportunities to exploit generative model for discriminative tasks.}

\hiddensection{Acknowledgement}
Weidi Xie would like to acknowledge the support of Yimeng Long and Yike Xie in enabling his contribution. 
This research was supported by the National Key R\&D Program of China (No.~2022ZD0161400).

{\small
\bibliographystyle{ieee_fullname}
\bibliography{egbib}

\begin{thebibliography}{10}\itemsep=-1pt

\bibitem{baek2021exploiting}
Donghyeon Baek, Youngmin Oh, and Bumsub Ham.
\newblock Exploiting a joint embedding space for generalized zero-shot semantic
  segmentation.
\newblock In {\em Proc. ICCV}, 2021.

\bibitem{brock2018large}
Andrew Brock, Jeff Donahue, and Karen Simonyan.
\newblock Large scale gan training for high fidelity natural image synthesis.
\newblock {\em arXiv preprint arXiv:1809.11096}, 2018.

\bibitem{bucher2019zero}
Maxime Bucher, Tuan-Hung Vu, Matthieu Cord, and Patrick P{\'e}rez.
\newblock Zero-shot semantic segmentation.
\newblock {\em NeurIPS}, 2019.

\bibitem{cheng2021per}
Bowen Cheng, Alex Schwing, and Alexander Kirillov.
\newblock Per-pixel classification is not all you need for semantic
  segmentation.
\newblock {\em NeurIPS}, 2021.

\bibitem{cheng2021sign}
Jiaxin Cheng, Soumyaroop Nandi, Prem Natarajan, and Wael Abd-Almageed.
\newblock Sign: Spatial-information incorporated generative network for
  generalized zero-shot semantic segmentation.
\newblock In {\em Proc. ICCV}, 2021.

\bibitem{dhariwal2021diffusion}
Prafulla Dhariwal and Alexander Nichol.
\newblock Diffusion models beat gans on image synthesis.
\newblock {\em NeurIPS}, 2021.

\bibitem{ding2022decoupling}
Jian Ding, Nan Xue, Gui-Song Xia, and Dengxin Dai.
\newblock Decoupling zero-shot semantic segmentation.
\newblock In {\em Proc. CVPR}, 2022.

\bibitem{dosovitskiy2020image}
Alexey Dosovitskiy, Lucas Beyer, Alexander Kolesnikov, Dirk Weissenborn,
  Xiaohua Zhai, Thomas Unterthiner, Mostafa Dehghani, Matthias Minderer, Georg
  Heigold, Sylvain Gelly, et~al.
\newblock An image is worth 16x16 words: Transformers for image recognition at
  scale.
\newblock {\em arXiv preprint arXiv:2010.11929}, 2020.

\bibitem{everingham2010pascal}
Mark Everingham, Luc Van~Gool, Christopher~KI Williams, John Winn, and Andrew
  Zisserman.
\newblock The pascal visual object classes (voc) challenge.
\newblock {\em International journal of computer vision(IJCV)}, 2010.

\bibitem{goodfellow2020generative}
Ian Goodfellow, Jean Pouget-Abadie, Mehdi Mirza, Bing Xu, David Warde-Farley,
  Sherjil Ozair, Aaron Courville, and Yoshua Bengio.
\newblock Generative adversarial networks.
\newblock {\em Communications of the ACM}, 2020.

\bibitem{gu2020context}
Zhangxuan Gu, Siyuan Zhou, Li Niu, Zihan Zhao, and Liqing Zhang.
\newblock Context-aware feature generation for zero-shot semantic segmentation.
\newblock In {\em ACM MM}, 2020.

\bibitem{ho2020denoising}
Jonathan Ho, Ajay Jain, and Pieter Abbeel.
\newblock Denoising diffusion probabilistic models.
\newblock {\em NeurIPS}, 2020.

\bibitem{ho2022cascaded}
Jonathan Ho, Chitwan Saharia, William Chan, David~J Fleet, Mohammad Norouzi,
  and Tim Salimans.
\newblock Cascaded diffusion models for high fidelity image generation.
\newblock {\em J. Mach. Learn. Res.}, 2022.

\bibitem{hong2019learning}
Richang Hong, Daqing Liu, Xiaoyu Mo, Xiangnan He, and Hanwang Zhang.
\newblock Learning to compose and reason with language tree structures for
  visual grounding.
\newblock {\em IEEE transactions on pattern analysis and machine
  intelligence(TPAMI)}, 2019.

\bibitem{kingma2014adam}
Diederik~P Kingma and Jimmy Ba.
\newblock Adam: A method for stochastic optimization.
\newblock {\em arXiv preprint arXiv:1412.6980}, 2014.

\bibitem{kingma2018glow}
Durk~P Kingma and Prafulla Dhariwal.
\newblock Glow: Generative flow with invertible 1x1 convolutions.
\newblock {\em NeurIPS}, 2018.

\bibitem{kingma2013auto}
Diederik~P Kingma and Max Welling.
\newblock Auto-encoding variational bayes.
\newblock {\em arXiv preprint arXiv:1312.6114}, 2013.

\bibitem{li2022bigdatasetgan}
Daiqing Li, Huan Ling, Seung~Wook Kim, Karsten Kreis, Sanja Fidler, and Antonio
  Torralba.
\newblock Bigdatasetgan: Synthesizing imagenet with pixel-wise annotations.
\newblock In {\em Proc. CVPR}, 2022.

\bibitem{liao2020real}
Yue Liao, Si Liu, Guanbin Li, Fei Wang, Yanjie Chen, Chen Qian, and Bo Li.
\newblock A real-time cross-modality correlation filtering method for referring
  expression comprehension.
\newblock In {\em Proc. CVPR}, 2020.

\bibitem{lin2014microsoft}
Tsung-Yi Lin, Michael Maire, Serge Belongie, James Hays, Pietro Perona, Deva
  Ramanan, Piotr Doll{\'a}r, and C~Lawrence Zitnick.
\newblock Microsoft coco: Common objects in context.
\newblock In {\em Proc. ECCV}, 2014.

\bibitem{liu2019learning}
Daqing Liu, Hanwang Zhang, Feng Wu, and Zheng-Jun Zha.
\newblock Learning to assemble neural module tree networks for visual
  grounding.
\newblock In {\em Proc. ICCV}, 2019.

\bibitem{liu2021swin}
Ze Liu, Yutong Lin, Yue Cao, Han Hu, Yixuan Wei, Zheng Zhang, Stephen Lin, and
  Baining Guo.
\newblock Swin transformer: Hierarchical vision transformer using shifted
  windows.
\newblock In {\em Proc. ICCV}, 2021.

\bibitem{nichol2021improved}
Alexander~Quinn Nichol and Prafulla Dhariwal.
\newblock Improved denoising diffusion probabilistic models.
\newblock In {\em ICML}, 2021.

\bibitem{pastore2021closer}
Giuseppe Pastore, Fabio Cermelli, Yongqin Xian, Massimiliano Mancini, Zeynep
  Akata, and Barbara Caputo.
\newblock A closer look at self-training for zero-label semantic segmentation.
\newblock In {\em Proc. CVPRW}, 2021.

\bibitem{radford2021learning}
Alec Radford, Jong~Wook Kim, Chris Hallacy, Aditya Ramesh, Gabriel Goh,
  Sandhini Agarwal, Girish Sastry, Amanda Askell, Pamela Mishkin, Jack Clark,
  et~al.
\newblock Learning transferable visual models from natural language
  supervision.
\newblock In {\em ICML}, 2021.

\bibitem{ramesh2022hierarchical}
Aditya Ramesh, Prafulla Dhariwal, Alex Nichol, Casey Chu, and Mark Chen.
\newblock Hierarchical text-conditional image generation with clip latents.
\newblock {\em arXiv preprint arXiv:2204.06125}, 2022.

\bibitem{rombach2022high}
Robin Rombach, Andreas Blattmann, Dominik Lorenz, Patrick Esser, and Bj{\"o}rn
  Ommer.
\newblock High-resolution image synthesis with latent diffusion models.
\newblock In {\em Proc. CVPR}, 2022.

\bibitem{sadhu2019zero}
Arka Sadhu, Kan Chen, and Ram Nevatia.
\newblock Zero-shot grounding of objects from natural language queries.
\newblock In {\em Proc. ICCV}, 2019.

\bibitem{saharia2022palette}
Chitwan Saharia, William Chan, Huiwen Chang, Chris Lee, Jonathan Ho, Tim
  Salimans, David Fleet, and Mohammad Norouzi.
\newblock Palette: Image-to-image diffusion models.
\newblock In {\em ACM SIGGRAPH}, 2022.

\bibitem{saharia2022photorealistic}
Chitwan Saharia, William Chan, Saurabh Saxena, Lala Li, Jay Whang, Emily
  Denton, Seyed Kamyar~Seyed Ghasemipour, Burcu~Karagol Ayan, S~Sara Mahdavi,
  Rapha~Gontijo Lopes, et~al.
\newblock Photorealistic text-to-image diffusion models with deep language
  understanding.
\newblock {\em NeurIPS}, 2022.

\bibitem{saharia2022image}
Chitwan Saharia, Jonathan Ho, William Chan, Tim Salimans, David~J Fleet, and
  Mohammad Norouzi.
\newblock Image super-resolution via iterative refinement.
\newblock {\em IEEE transactions on pattern analysis and machine
  intelligence(TPAMI)}, 2022.

\bibitem{song2020denoising}
Jiaming Song, Chenlin Meng, and Stefano Ermon.
\newblock Denoising diffusion implicit models.
\newblock {\em arXiv preprint arXiv:2010.02502}, 2020.

\bibitem{tang2022daam}
Raphael Tang, Akshat Pandey, Zhiying Jiang, Gefei Yang, Karun Kumar, Jimmy Lin,
  and Ferhan Ture.
\newblock What the daam: Interpreting stable diffusion using cross attention.
\newblock {\em arXiv preprint arXiv:2210.04885}, 2022.

\bibitem{van2016conditional}
Aaron Van~den Oord, Nal Kalchbrenner, Lasse Espeholt, Oriol Vinyals, Alex
  Graves, et~al.
\newblock Conditional image generation with pixelcnn decoders.
\newblock {\em NeurIPS}, 2016.

\bibitem{wang2018learning}
Liwei Wang, Yin Li, Jing Huang, and Svetlana Lazebnik.
\newblock Learning two-branch neural networks for image-text matching tasks.
\newblock {\em IEEE Transactions on Pattern Analysis and Machine
  Intelligence(TPAMI)}, 2018.

\bibitem{wang2019neighbourhood}
Peng Wang, Qi Wu, Jiewei Cao, Chunhua Shen, Lianli Gao, and Anton van~den
  Hengel.
\newblock Neighbourhood watch: Referring expression comprehension via
  language-guided graph attention networks.
\newblock In {\em Proc. CVPR}, 2019.

\bibitem{xian2019semantic}
Yongqin Xian, Subhabrata Choudhury, Yang He, Bernt Schiele, and Zeynep Akata.
\newblock Semantic projection network for zero-and few-label semantic
  segmentation.
\newblock In {\em Proc. CVPR}, 2019.

\bibitem{yang2019dynamic}
Sibei Yang, Guanbin Li, and Yizhou Yu.
\newblock Dynamic graph attention for referring expression comprehension.
\newblock In {\em Proc. ICCV}, 2019.

\bibitem{yang2020improving}
Zhengyuan Yang, Tianlang Chen, Liwei Wang, and Jiebo Luo.
\newblock Improving one-stage visual grounding by recursive sub-query
  construction.
\newblock In {\em Proc. ECCV}, 2020.

\bibitem{yang2019fast}
Zhengyuan Yang, Boqing Gong, Liwei Wang, Wenbing Huang, Dong Yu, and Jiebo Luo.
\newblock A fast and accurate one-stage approach to visual grounding.
\newblock In {\em Proc. ICCV}, 2019.

\bibitem{yu2022scaling}
Jiahui Yu, Yuanzhong Xu, Jing~Yu Koh, Thang Luong, Gunjan Baid, Zirui Wang,
  Vijay Vasudevan, Alexander Ku, Yinfei Yang, Burcu~Karagol Ayan, et~al.
\newblock Scaling autoregressive models for content-rich text-to-image
  generation.
\newblock {\em arXiv preprint arXiv:2206.10789}, 2022.

\end{thebibliography}
}

\clearpage
\onecolumn
\appendix
\renewcommand*\contentsname{Supplementary}
\tableofcontents
\newpage
In this supplementary document, 
we start by giving more details on the architecture of our grounding module in Section~\ref{section 1}, followed by details for generating the training dataset in Section~\ref{section 2}; then describe the additional ablation studies in Section~\ref{section 3}, as promised in the main paper; 
In Section~\ref{section 5}, we present additional qualitative results; Lastly, we illustrate the limitation of our method and future work in Section~\ref{section 6}.

\section{Details on the Architecture of Grounding module}
\label{section 1}
\begin{figure}[ht]
    \centering
    \includegraphics[width=\linewidth]{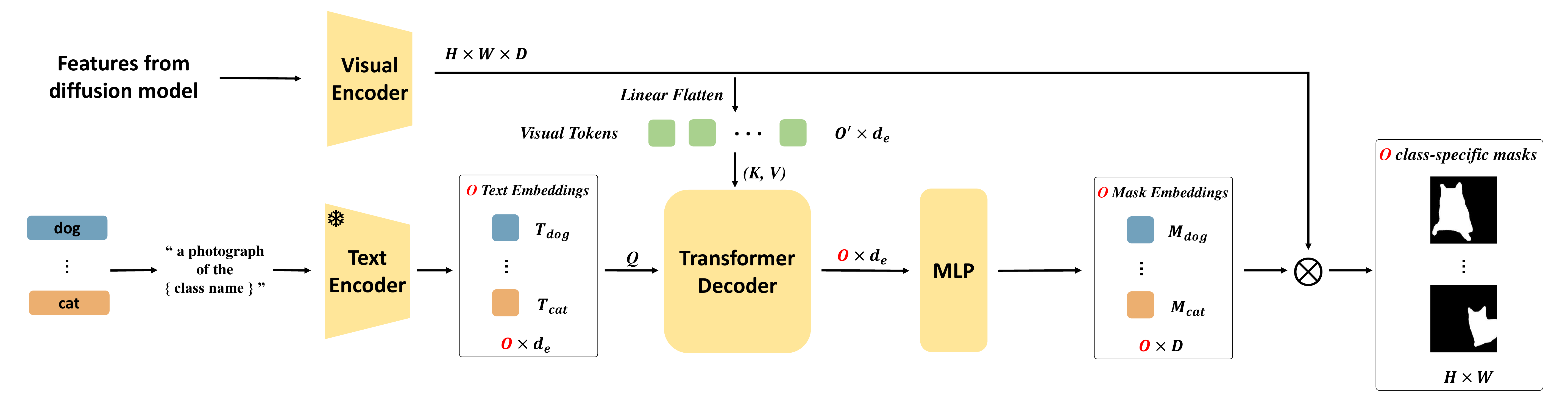}
    \vspace{-0.7cm}
    \caption{\textbf{Detailed architecture of our grounding module.} 
    We first generate $\mathcal{O}$ text embeddings by injecting the class names into a prompt template and then feed them to a pre-trained text encoder. 
    The visual encoder takes the features from Stable Diffusion as input and outputs fused visual features, which are then flattened to a sequence of visual tokens. Next, we feed the visual tokens into a transformer decoder as \textit{Key} and \textit{Value}, and feed text embeddings as \textit{Query}. The outputs of transformer decoder are then fed into an MLP to obtain $\mathcal{O}$ mask embeddings. Mask embeddings are dot producted with the output features of visual encoder to generate $\mathcal{O}$ class-specific binary masks.
    }
    \label{fig:framework_supp}
    \vspace{-0.2cm}
\end{figure}

We show the detailed architecture of our grounding module in Fig.~\ref{fig:framework_supp}, which consists of visual encoder, text encoder, transformer decoder and MLP in the fusion module.

\noindent \textbf{Visual Encoder.}
The input $\{f^1,\dots ,f^n\}$ are extracted from Stable Diffusion~\cite{rombach2022high},
and the visual encoder aims to upsample and fuse the visual feature,
and output the visual feature map, $\hat{\mathcal{F}} = \Phi_{\text{v-enc}}(\{f^1, \dots, f^n\}), \hat{\mathcal{F}} \in \mathbb{R}^{H \times W \times D}$ (here we use $H = W = 512$, and $D = 240$). Note that, the resolution of the fused visual feature is the same as the resolution of the generated image, and the segmentation masks.

\vspace{5pt}
\noindent \textbf{Text Encoder.}
We adopt the pre-trained text encoder from CLIP~\cite{radford2021learning}, 
which is also used in Stable Diffusion~\cite{rombach2022high}. 
It takes text prompt $y$ as input and outputs the corresponding text embedding: $\mathcal{E}_{\text{object}}=\Phi_{\text{t-enc}}(y), \mathcal{E}_{\text{object}} \in \mathbb{R}^{\mathcal{O} \times d_{\text{text}}}$, where $\mathcal{O}$ is the total number of objects of interest and $d_{\text{text}} = 768$.

\vspace{5pt}
\noindent \textbf{Transformer Decoder in Fusion Module.}
Similar to the operation in standard ViT architecture~\cite{dosovitskiy2020image}, 
we convert the visual features $\hat{\mathcal{F}} \in \mathbb{R}^{H \times W \times D}$ into visual tokens $\hat{\mathcal{F}}_{\text{flatten}} \in \mathbb{R}^{\mathcal{\mathcal{O}}' \times d_{\text{visual}}}$, 
where $\mathcal{O}' = \frac{HW}{p^2} = 16384$ is the number of tokens, 
$p$ refers to the patch size and $d_{\text{visual}} = p^2 \times D = 3840$. 
Then the visual tokens and text embeddings are mapped into the same dimension $d_{e} = 512$ with MLPs, and passed into the  transformer decoder with three layers.
The text embeddings are treated as \texttt{query} with dimension $\mathcal{O} \times d_e$, and the visual tokens are treated as \texttt{key} and \texttt{value} with dimension $\mathcal{O}' \times d_e$.
The output of transformer decoder is of the same resolution as \texttt{query}, with dimension $\mathcal{O} \times d_e$.

\vspace{4pt}
\noindent \textbf{MLP in Fusion Module.}
At last, we use an MLP to map the output of transformer decoder into mask embeddings with dimension $\mathcal{O} \times D$, which are then dot producted with the fused visual feature ($\hat{\mathcal{F}} \in \mathbb{R}^{H \times W \times D}$) to generate class-specific binary masks ($\mathcal{O} \times H \times W$), each mask is of the same spatial resolution of the generated image.

\newpage

\section{Details on the Synthetic Dataset}
\label{section 2}
\subsection{Dataset Split}

Here, to train our proposed grounding module, and properly evaluate its ability for segmenting the objects that are unseen at training time, 
we construct the training dataset with images of only seen categories, 
and the test dataset consists of both seen and unseen categories.
The detail of the split on PASCAL-sim and COCO-sim, 
{\em i.e.} the split of seen categories and unseen categories, is shown in Tab.~\ref{tab:categories-split}, where PASCAL-sim has 15 seen categories and 5 unseen categories, COCO-sim has 65 seen categories and 14 unseen categories. 
Note that, we ignore the category: `mouse' in the COCO-sim since the diffusion model generates `rat' in the image for the category: `mouse', 
while  `mouse' in the vocabulary of the off-the-shelf detector means mouse as a computer accessory, thus the detector fails to detect the category `mouse' in the  image generated by the diffusion model.

\begin{table}[h]
\footnotesize
\begin{center}
\tabcolsep=0.1cm
\begin{tabular}{c|c|p{10.5cm}|p{3cm}}
\cmidrule( ){1-4}
& \multicolumn{1}{c|}{Categories} & \multicolumn{1}{c|}{seen} & \multicolumn{1}{c}{Unseen}\\
\cmidrule( ){1-4}
\multirow{6}{*}{\textbf{PASCAL-sim}}& $\mathrm{Split} 1$ &  aeroplane, bicycle, bird,  boat, bottle, bus, cat, chair, cow, diningtable, horse, motorbike, person, pottedplant,  sheep  &  tvmonitor, car, dog, sofa, train \\
\cmidrule( ){2-4}
& $\mathrm{Split} 2$ & tvmonitor, car, dog, sofa, train,  aeroplane,  bicycle,  bird,  boat,  bottle,  bus,  cat,  chair,  cow,  diningtable   &  horse,  motorbike,  person,  pottedplant,  sheep \\
\cmidrule( ){2-4}
& $\mathrm{Split} 3$ &  horse,  motorbike,  person,  pottedplant,  sheep,  tvmonitor,  car,  dog,  sofa,  train,  aeroplane,  bicycle,  bird,  boat,  bottle  &  bus,  cat,  chair,  cow,  diningtable \\
\cmidrule( ){1-4}
\multirow{18}{*}{\textbf{COCO-sim}}& $\mathrm{Split} 1$ &  person, bicycle, car, motorbike, bus, truck, boat, traffic light, fire hydrant, stop sign,  bench, bird, dog, horse, sheep, cow, elephant, zebra, giraffe,  backpack, umbrella, handbag, tie, skis, sports ball, kite, baseball bat, baseball glove,  skateboard, surfboard, tennis racket,  bottle, wine glass, cup, knife, spoon, bowl, banana, apple, orange, broccoli, carrot,  pizza, donut, cake, chair, bench, pottedplant, bed, diningtable, tvmonitor, laptop, remote, keyboard, cell phone,  microwave, oven, sink, refrigerator, book, clock, vase, scissors, teddy bear, toothbrush    &   aeroplane, train, parking meter, cat, bear, suitcase, frisbee, snowboard, fork, sandwich, hot dog, toilet, toaster, hair drier \\
\cmidrule( ){2-4}
& $\mathrm{Split} 2$ & aeroplane, train, parking meter, cat, bear, suitcase, frisbee, snowboard, fork, sandwich, hot dog, toilet, toaster, hair drier, person, bicycle, car, motorbike, bus, truck, boat, traffic light, fire hydrant, stop sign, bench, bird, dog, horse, sheep, cow, elephant, zebra, giraffe, backpack, umbrella, handbag, tie, skis, sports ball, kite, baseball bat, baseball glove, skateboard, surfboard, tennis racket, bottle, wine glass, cup, knife,  spoon, bowl, banana, apple, orange, broccoli, carrot, pizza, donut, cake, chair, bench, pottedplant, bed, diningtable, tvmonitor   &  laptop, remote, keyboard, cell phone, microwave, oven, sink, refrigerator,  book, clock, vase, scissors, teddy bear, toothbrush \\
\cmidrule( ){2-4}
& $\mathrm{Split} 3$ &  laptop, remote, keyboard, cell phone, microwave, oven, sink, refrigerator, book, clock, vase, scissors, teddy bear, toothbrush, aeroplane, train, parking meter, cat, bear, suitcase, frisbee, snowboard, fork, sandwich, hot dog, toilet, toaster, hair drier, person, bicycle, car, motorbike, bus, truck, boat, traffic light, fire hydrant, stop sign, bench, bird, dog, horse, sheep, cow, elephant, zebra, giraffe, backpack, umbrella, handbag,  tie, skis, sports ball, kite, baseball bat, baseball glove, skateboard, surfboard, tennis racket, bottle,  wine glass, cup, knife, spoon, bowl    &   banana, apple, orange, broccoli, carrot, pizza, donut, cake, chair, bench, pottedplant, bed, diningtable, tvmonitor  \\
\cmidrule( ){1-4}
\end{tabular}
\end{center}
\vspace{-0.6cm}
\caption{\textbf{The details on the split of categories on PASCAL-sim and COCO-sim.} }
\label{tab:categories-split}
\vspace{-0.1cm}
\end{table}


\newpage

\subsection{Dataset for Training Grounding Module}
\label{section:Grounded Dataset}

To construct the training set, 
(1) we first randomly select one or two categories from the seen ones, 
where objects tend to co-appear in natural images, 
based on the annotation in PASCAL VOC~\cite{everingham2010pascal} or COCO~\cite{lin2014microsoft}, called \textbf{co-appearing category pair}, 
and use the prompt template to decorate these selected categories, 
thus we can obtain the text prompt; 
(2) we pass the text prompt and randomly sampled Gaussian noise to the Stable Diffusion~\cite{rombach2022high} to obtain the generated image; 
(3) next, we pass the generated image to the off-the-shelf detector to obtain the oracle segmentation mask; 
(4) finally, we can construct the triplet which consists of the generated image, oracle segmentation mask, and text prompt; 
(5) repeat the above procedure, we can generate infinite triplets for the training set. Algorithm~\ref{alg:pseudo1} displays the procedure for generating the training set. 

To construct the test set for evaluating the grounding module, 
we can use a procedure similar to the training set. 
The differences are: 
(i) we use all categories, including seen and unseen categories to construct the test set. 
(ii) to obtain more reliable test results, 
we only add the triplet to the test set when the generated image and oracle segmentation mask have high quality which is checked manually, 
{\em i.e.}, the generated image by Stable Diffusion contains the recognizable objects of selected categories, and the off-the-shelf detector successfully produces the high-quality oracle segmentation mask. 

In this paper, PASCAL-sim has 20 categories and 142 co-appearing category pairs. We construct 30 triplets per category and 5 triplets per co-appearing category pair for PASCAL-sim test set. In total, PASCAL-sim test set has 1310 triplets. 
COCO-sim has 79 categories and 1559 co-appearing category pairs. 
We construct 30 triplets per category and 2 triplets per co-appearing category pair for COCO-sim test set. In total, COCO-sim test set has 5488 triplets.

\begin{algorithm}[!htb]
\caption{Constructing the dataset for training grounding module~(pseudocode in PyTorch-like style).}
 \label{alg:pseudo1}
 \definecolor{codeblue}{rgb}{0.25,0.5,0.5}
\lstset{
 backgroundcolor=\color{white},
 basicstyle=\fontsize{7.2pt}{7.2pt}\ttfamily\selectfont,
 columns=fullflexible,
 breaklines=true,
 captionpos=b,
 escapeinside={(:}{:)},
 commentstyle=\fontsize{7.2pt}{7.2pt}\color{codeblue},
 keywordstyle=\fontsize{7.2pt}{7.2pt},
 }
\begin{lstlisting}[language=python]
# C_seen: the list of seen categories
# img_shape: the shape of expected generated image
# exp_train_size: the expected size of  training set
# n: the number of selected categories, n = 1 or 2
# co-appearing_category_pair_list: a list containing all co-appearing category pairs, where objects tend to 
# co-appear in natural images, based on the annotation in PASCAL VOC or COCO

D_train = [] #initialize the training set

while (len(D_train) < exp_train_size):

    y = None #initialize the text prompt
    
    #randomly select n categories from seen categories
    selected_class_list = random_select(C_seen, n) 
    
    if n = 1:
        class = select_class_list[0]
        
        # decorate the selected category by a pre-defined prompt template, e.g., "a photograph of a [class name]"
        y = prompt_template(class) 
        
    else if n = 2:
        class1, class2 = select_class_list[0], select_class_list[1]
        if (class1, class2) in co-appearing_category_pair_list:
        
            # decorate the selected categories by a pre-defined prompt template, e.g., "a photograph of a 
            # [class1 name] and a [class2 name]"
            y = prompt_template(class1, class2) 
            
    if y != None:
        
        #randomly sample a Gaussian noise epsilon
        epsilon=torch.randn(img_shape)
        
        # pass the noise and text prompt to the diffusion model to generate image I
        I = diffusion_model(epsilon, y) 
        
        # pass the generated image to the off-the-shelf detector to obtain the oracle segmentation mask m
        m = pretrain_detector(I) 
        
        # add the triplet (generated image, oracle segmentation mask, text prompt) to the training set
        D_train.append((I, m, y)) 
\end{lstlisting}
\end{algorithm}



\newpage

\subsection{Dataset for Training Semantic Segmentation Model}
As discussed in the main text, 
we synthesize two semantic segmentation datasets for all 20 categories in PASCAL VOC~\cite{everingham2010pascal} and 79 categories in COCO~\cite{lin2014microsoft}, respectively. 
We first randomly select one or two categories~(co-appearing category pair), 
to construct the text prompt, 
and then pass randomly sampled Gaussian noise and text prompt to the diffusion model to obtain the generated image,
and use our proposed grounding module to get the corresponding segmentation mask. Thus, we can get the pair of generated image and segmentation mask. Repeat the above procedure, we can obtain the synthetic semantic segmentation datasets at a large scale. Algorithm~\ref{alg:pseudo3} displays the procedure for generating the synthetic semantic segmentation dataset.

In this paper, for categories in PASCAL VOC, the synthetic semantic segmentation dataset consists of 500 images per category and 71 images per co-appearing category pair. Thus, there exist 10k images for 20 categories and 10082 images for 142 co-appearing category pairs in total. For categories in COCO, the dataset consists of 1500 images per category and 70 images per co-appearing category pair, thus there exist 118500 images for 79 categories and 109130 images for 1559 co-appearing category pairs

\begin{algorithm}[!htb]
\caption{Pseudo-code for generating the synthetic semantic segmentation dataset in a PyTorch-like style.}
 \label{alg:pseudo3}
 \definecolor{codeblue}{rgb}{0.25,0.5,0.5}
\lstset{
 backgroundcolor=\color{white},
 basicstyle=\fontsize{7.2pt}{7.2pt}\ttfamily\selectfont,
 columns=fullflexible,
 breaklines=true,
 captionpos=b,
 escapeinside={(:}{:)},
 commentstyle=\fontsize{7.2pt}{7.2pt}\color{codeblue},
 keywordstyle=\fontsize{7.2pt}{7.2pt},
 }
\begin{lstlisting}[language=python]
# C: the list of  all categories
# img_shape: the shape of expected generated image
# exp_dataset_size: the expected size of synthesis semantic segmentation dataset
# n: the number of selected categories, n = 1 or 2
# co-appearing_category_pair_list: a list containing all co-appearing category pairs, where objects tend to 
# co-appear in natural images, based on the annotation in PASCAL VOC or COCO

D_seg = [] #initialize the synthesis semantic segmentation dataset

while (len(D_seg) < exp_dataset_size):

    y = None #initialize the text prompt
    
    #randomly select n categories from all categories
    selected_class_list = random_select(C, n) 
    
    if n = 1:
        class = select_class_list[0]
        
        # decorate the selected category by a pre-defined prompt template, e.g., "a photograph of a [class name]"
        y = prompt_template(class) 
        
    else if n = 2:
        class1, class2 = select_class_list[0], select_class_list[1]
        if (class1, class2) in co-appearing_category_pair_list:
        
            # decorate the selected categories by a pre-defined prompt template, e.g., "a photograph of a 
            # [class1 name] and a [class2 name]"
            y = prompt_template(class1, class2) 
            
    if y != None:
        
        #randomly sample a Gaussian noise epsilon
        epsilon=torch.randn(img_shape)
        
        # pass the noise and text prompt to the diffusion model with grounding module to generate image I
        # and segmentaion mask m
        I, m = diffusion_model_with_grounding(epsilon, y) 
        
        # add the pair (generated image, generated segmentation mask) to the synthesis semantic segmentation dataset
        D_seg.append((I, m)) 
\end{lstlisting}
\end{algorithm}


\clearpage

\section{Additional Ablation Study}
\label{section 3}

\vspace{2pt}
\subsection{Synthetic Dataset Construction.}
We explore the effect of constructing different datasets for training the grounding module, by varying the number of objects in the images. 
As shown in Tab.~\ref{tab:ablation-2-final-testset}, training on the combination of one and two object categories gives the best results overall.

\begin{table}[ht]
\small
\begin{center}
\tabcolsep=0.12cm
\begin{tabular}{ccccccc}
\cmidrule( ){1-6}
 \multirow{3}{*}{\makecell[c]{Train Set \\ \# Objects  }}  & \multicolumn{2}{c}{One}  & \multicolumn{3}{c}{Two} \\
 \cmidrule(lr){2-3}\cmidrule(lr){4-6}
  & \makecell[c]{Seen} & \multicolumn{1}{c}{Unseen}
  & \makecell[c]{Seen} & \makecell[c]{Seen +Unseen} & \multicolumn{1}{c}{Unseen} \\
\cmidrule( ){1-6}
 single & \textbf{90.37}&\multicolumn{1}{c}{\textbf{83.85}} &43.89 &42.33 &38.91  \\
 two &88.35 &\multicolumn{1}{c}{82.93} &\textbf{80.56} &\textbf{68.08} & 56.36  \\
 mixture & 90.16 &\multicolumn{1}{c}{83.19} &78.93 & 66.07& \textbf{57.93} \\
                         
\cmidrule( ){1-6}
\end{tabular}
\end{center}
\vspace{-0.6cm}
\caption{\textbf{Ablation on dataset construction on PASCAL-sim.} The bolded number indicates the best result. Our model achieves the best performance when training on the combination of one and two object categories. }
\label{tab:ablation-2-final-testset}
\vspace{-0.45cm}
\end{table}






\vspace{5pt}
\subsection{Timesteps for Extracting Visual Representation.}
\begin{figure}[h]
    \centering
    \includegraphics[width=0.5\linewidth]{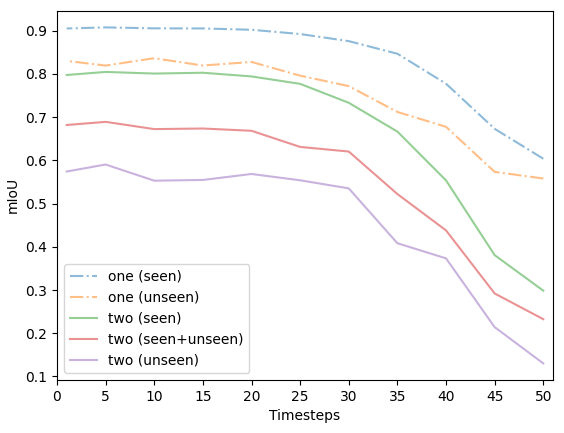}
    \vspace{-0.4cm}
    \caption{\textbf{Ablation on timesteps.} The mIoU is measured for the model with extracting features from Stable Diffusion in different timesteps on \textbf{PASCAL-sim}.}
    \label{fig:ablation-timestep}
    \vspace{-0.1cm}
\end{figure}

We compare the performance by extracting visual representation from Stable Diffusion at different timesteps, 
the results on PASCAL-sim can be seen in Fig.~\ref{fig:ablation-timestep}, 
showing that as the denoising steps gradually decrease, {\em i.e.}, from $t=0 \longrightarrow 50$, 
the performance for grounding tends to decrease in general, when $t=5$, the best result is obtained.

\vspace{5pt}
\subsection{Dataset Construction via DDIM Inverse.}
\label{sec:DDIMInverse}
In addition to using the off-the-shelf detectors, we also consider  constructing the training set by utilising the inverse process of diffusion to explicitly generate images close to those in the public dataset, for example, PASCAL VOC, and train the grounding module with the mask annotations available from the dataset.

Here, we describe an inverse procedure that enables to find a deterministic mapping from noise to images,
given the sampling rule being non-Markovian, for example, Denoising Diffusion Implicit Model~(DDIM)~\cite{song2020denoising} with the reverse process variance to be 0.
In DALL-E 2~\cite{ramesh2022hierarchical}, such inversion has been used to determine the noise that produces a specific image. In our considered Stable Diffusion, the image is first mapped to a latent vector $z^0$ by the pre-trained variational autoencoder (VAE), at each step of DDIM inversion,
$z^{t+1}$ is obtained from $z^t$ and the predicted noise term of U-Net, that takes $z^t$ and text prompt $y$ as input, ending up with an inverted noise $z^T$ eventually. 
In this paper, we exploit such DDIM inversion to train our grounding module with the dataset constructed from real image and segmentation masks. 

In particular, the first option enables to directly inherit the segmentation mask from the public dataset,
and the text prompt can be manually constructed by inserting class labels into the prompt template, 
for example, if the segmentation mask contains `dog' and `cat', 
the text prompt can be `a photograph of a dog and cat'. Besides, the visual feature can be obtained by extracting the feature from the U-Net of Stable Diffusion when $t=1$ at the inversion process.

\vspace{3pt}
\noindent \textbf{Constructed Dataset {\em v.s.}~Real Dataset.}
We explore the difference between training on constructed dataset and real dataset (PASCAL VOC) from two perspectives. 
{\em First}, we compare their performance on PASCAL-sim dataset for grounded generation in Tab.~\ref{tab:ablation-new}~(left). 
Though we successfully train our grounding module on real dataset, the domain gap limits its performance on grounded generation task. 
Considering PASCAL VOC only contains about 10k images, we adjust the construced dataset to the same magnitude and get better results. Additionally, due to the good scalability of the constructed dataset, the performance improves as the number of images increases. 
{\em Second}, we evaluate the grounding module on PASCAL VOC test dataset by DDIM inversion as shown in Tab.~\ref{tab:ablation-new}~(right).
Note that, under this circumstance, our model approximates a discriminative model. 
On seen categories of PASCAL VOC test set, the module trained on real dataset achieves the best result, while the module trained on constructed dataset gains an advantage on unseen categories.
Besides, we also train our module on both constructed dataset and real dataset, which results in great improvement on PASCAL VOC test dataset.

\begin{table}[h]
\footnotesize
\begin{center}
\tabcolsep=0.09cm
\begin{tabular}{cccccccc}
\cmidrule( ){1-8}
  \multirow{4}{*}{\makecell[c]{Dataset\\Type}}                        & \multicolumn{5}{c}{PASCAL-sim}                         & \multicolumn{2}{c}{\multirow{2}{*}{PASCAL-test}} \\
                         \cmidrule( ){2-6}
                         & \multicolumn{2}{c}{One} & \multicolumn{3}{c}{Two}      & \multicolumn{2}{c}{}                        \\
                         \cmidrule(lr){2-3}\cmidrule(lr){4-6}\cmidrule(lr){7-8}
                         & Seen       & Unseen     & Seen  & Seen+Unseen & Unseen & Seen                 & Unseen               \\
                         \cmidrule( ){1-8}
real                   & 75.67      & 61.26      & 64.08 & 49.23      & 45.14 & \textbf{75.19}                & 34.80                 \\
sim(10k)        & 88.77      & 70.04     & 73.07 & 57.08       & 46.59  & 61.75                & 48.42                \\ 
sim(40k)        & \textbf{90.16}      & \textbf{83.19}      & \textbf{78.93} & \textbf{66.07}       & \textbf{57.93}  & 64.44               & 53.86                \\
sim+real & 89.57     & 76.23      & 78.12 & 62.24       & 55.30  & 73.32                & \textbf{57.14}                \\


\cmidrule( ){1-8}
\end{tabular}
\end{center}
\vspace{-0.5cm}
\caption{\textbf{Ablation on the training dataset.} The bold numbers indicate the best results. Specifically, 'sim' and 'real' denote the constructed dataset and real dataset~(PASCAL-VOC), respectively. }
\label{tab:ablation-new}
\vspace{-0.2cm}
\end{table}


\begin{figure}[h]
    \centering
    \includegraphics[width=0.75\linewidth]{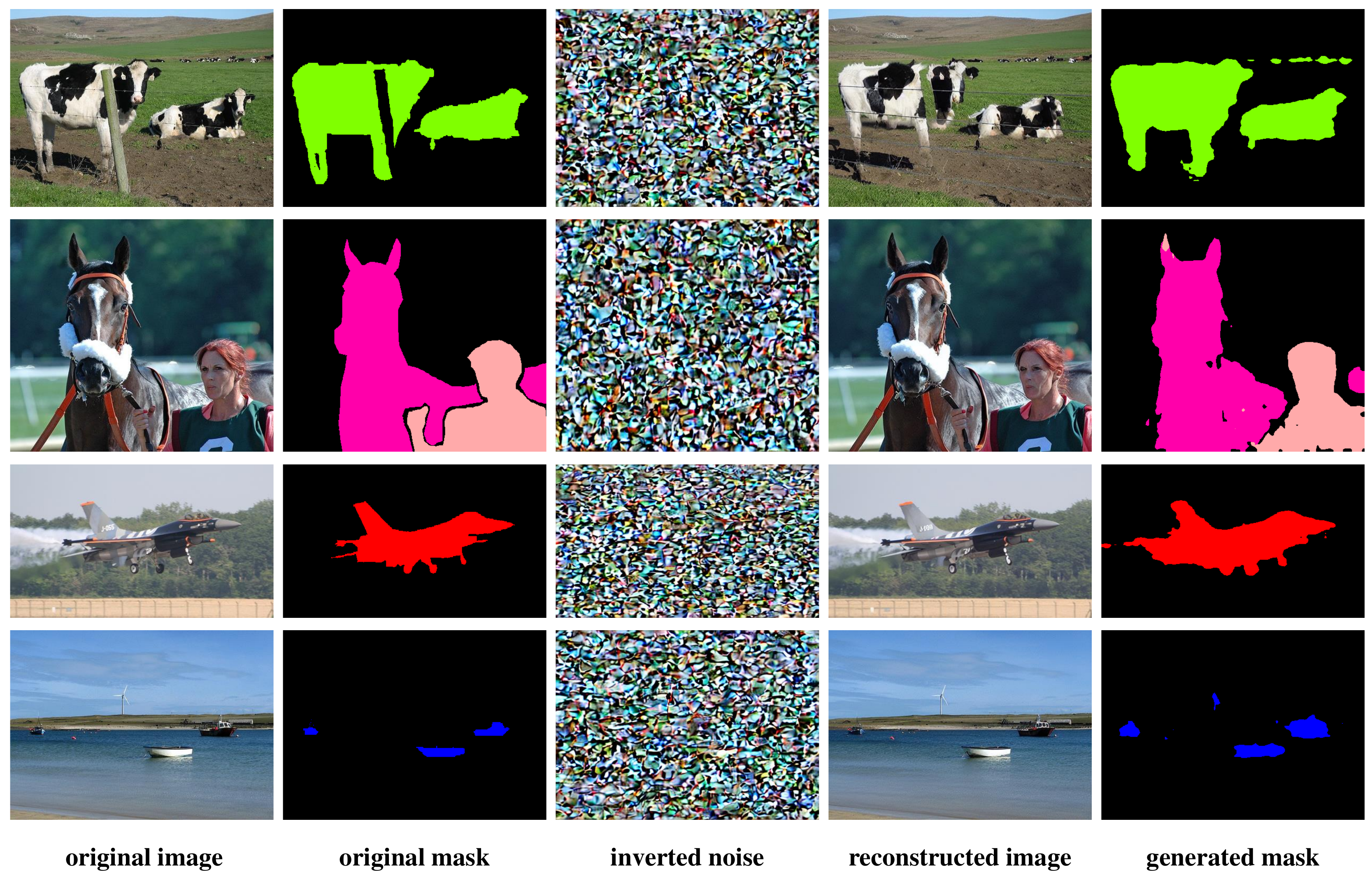}
    \vspace{-0.4cm}
    \caption{\textbf{Qualitative results on real dataset.} By DDIM Inversion, we firstly map the original images to corresponding noise, which are taken as input to regenerate images. And we train the grounding module to predict the masks during the regeneration process, with the supervision of the original masks. }
    \label{fig:ablation-ddiminverse}
    \vspace{-0.1cm}
\end{figure}

\newpage
\section{More Qualitative Results}
\label{section 5}
\begin{figure}[ht]
    \centering
    \includegraphics[width=\linewidth]{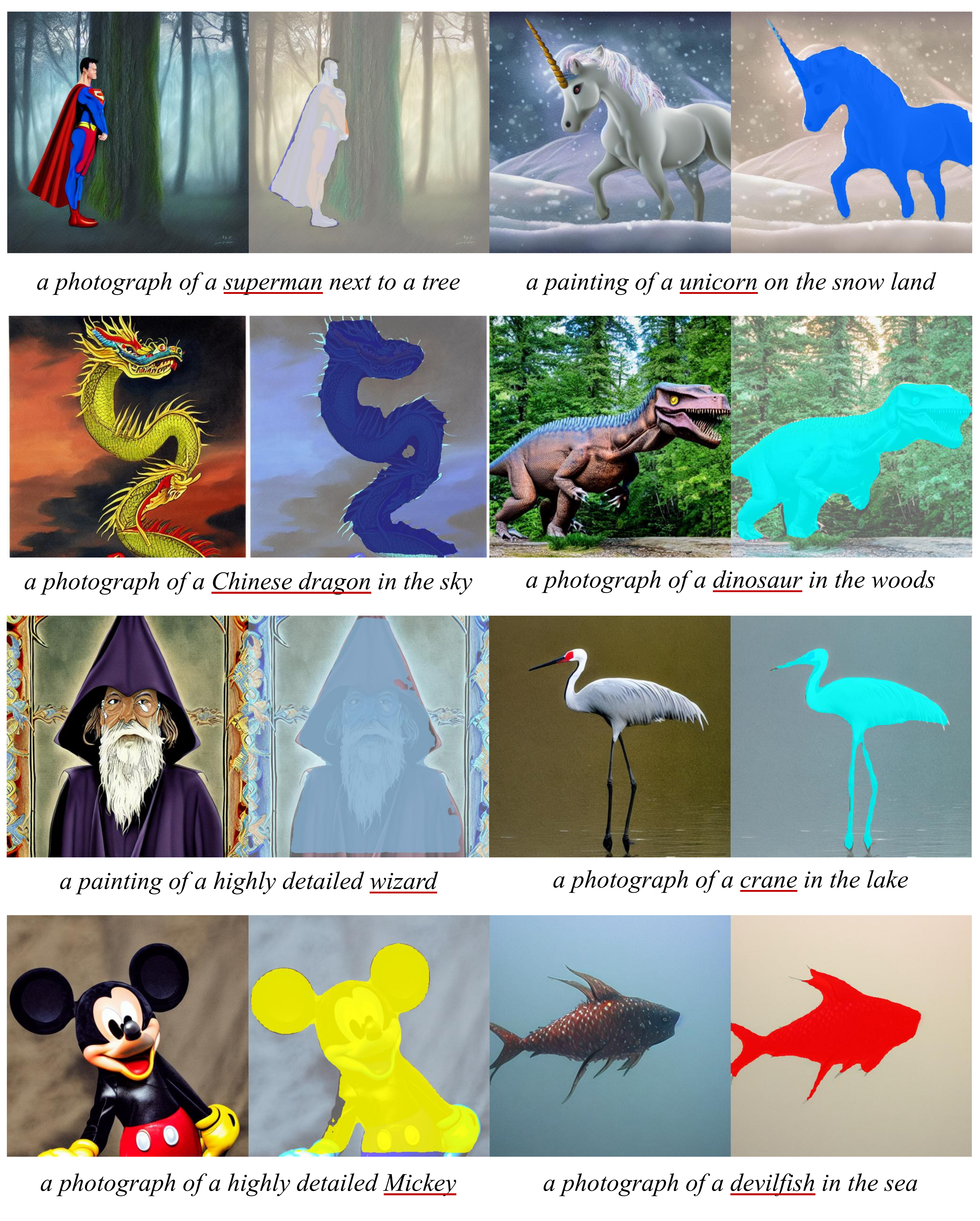}
    \vspace{-0.6cm}
    \caption{\textbf{Results of grounded generation.}
    The segmentation mask refers to the grounding results for the object underlined.
    }
    \label{fig:supp_2}
    \vspace{-0.2cm}
\end{figure}

\begin{figure}[ht]
    \centering
    \includegraphics[width=\linewidth]{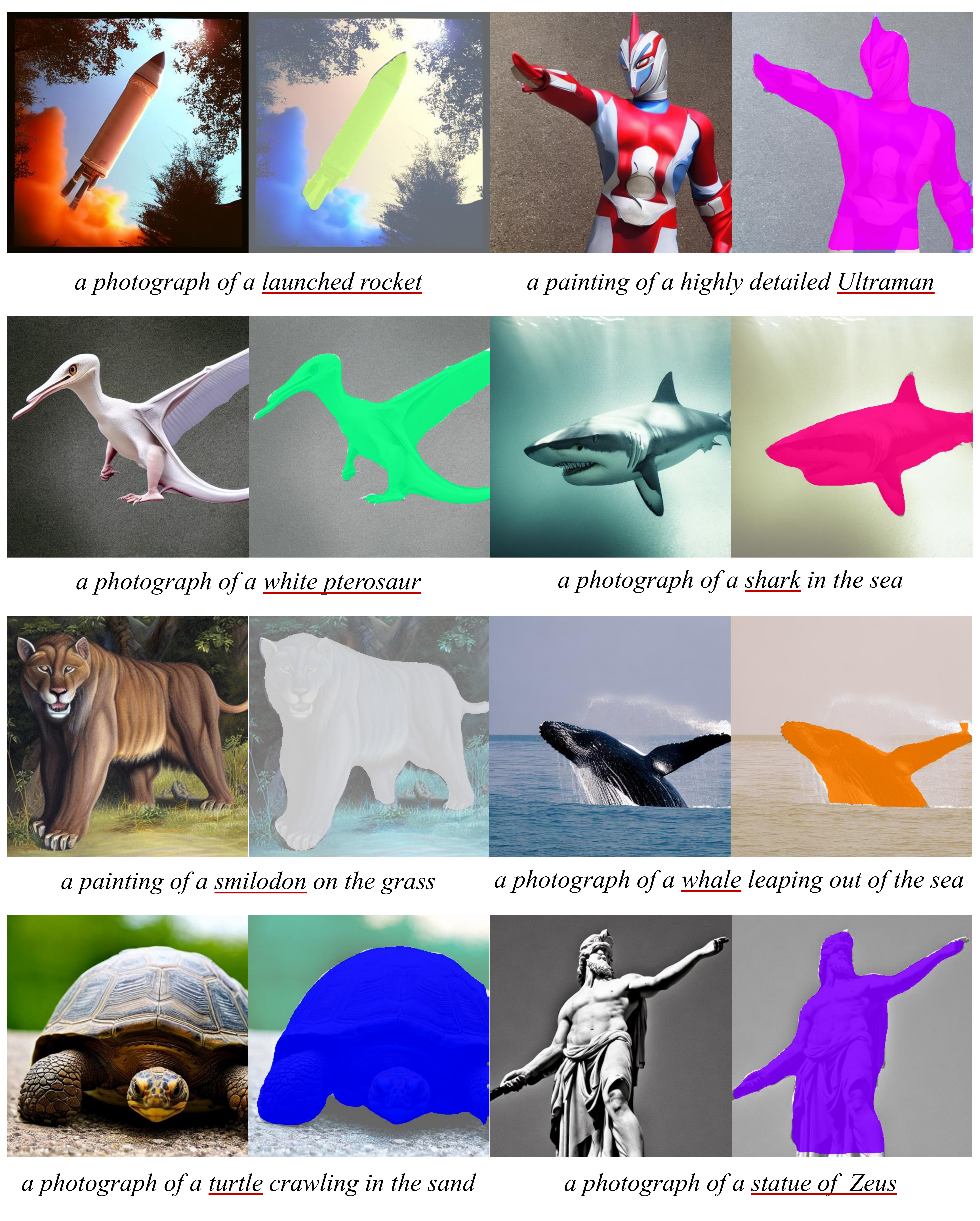}
    \vspace{-0.6cm}
    \caption{\textbf{Results of grounded generation.} 
    The segmentation mask refers to the grounding results for the object underlined.
    }
    \label{fig:supp_3}
    \vspace{-0.2cm}
\end{figure}

\begin{figure}[ht]
    \centering
    \includegraphics[width=\linewidth]{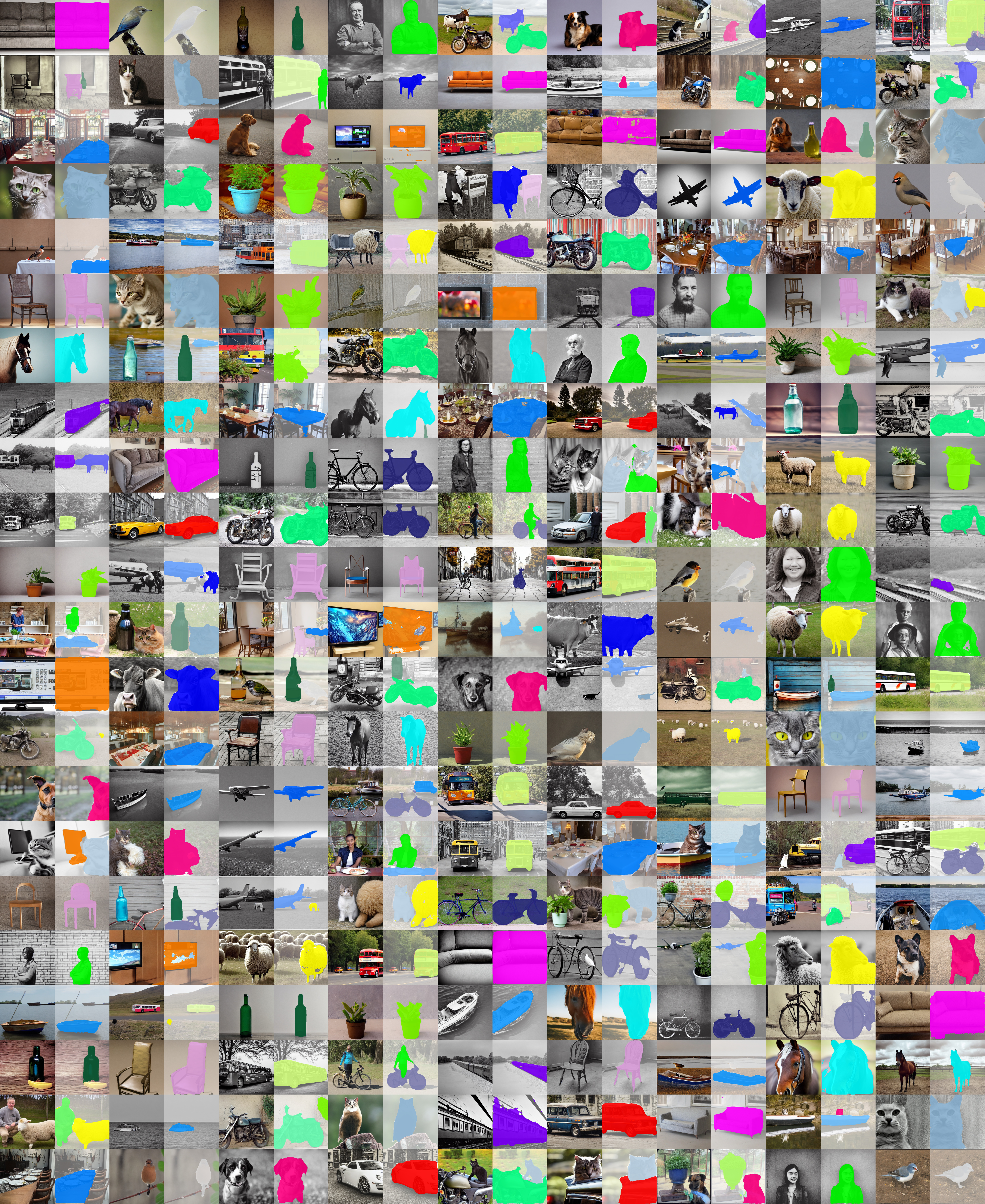}
    \vspace{-0.6cm}
    \caption{\textbf{Examples from our synthetic semantic segmentation dataset.}
    }
    \label{fig:result-supp}
    \vspace{-0.2cm}
\end{figure}

\begin{figure}[ht]
    \centering
    \includegraphics[width=\linewidth]{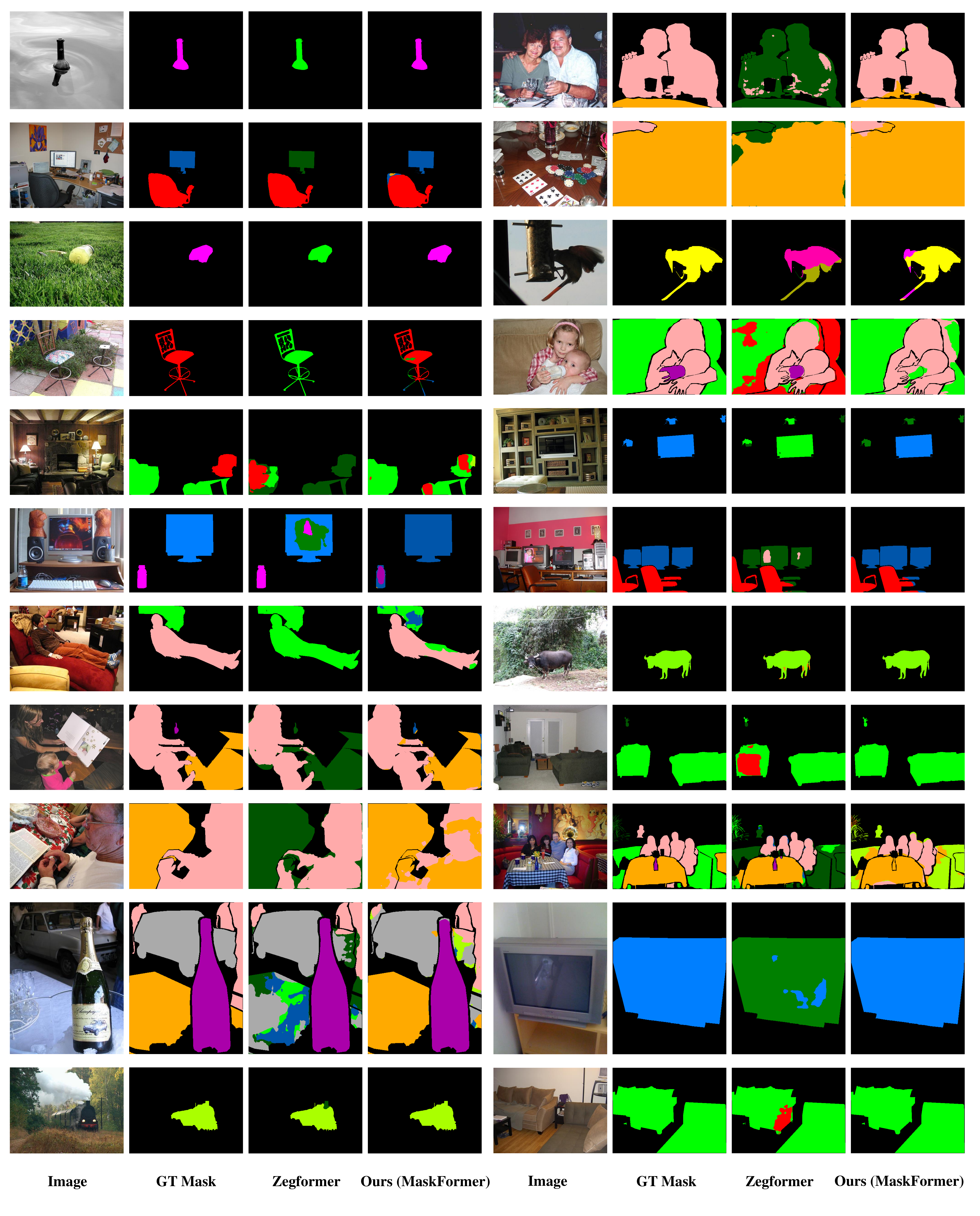}
    \vspace{-0.6cm}
    \caption{\textbf{More visualization of zero-shot segmentation results on Pascal VOC.} 
    }
    \label{fig:supp_5}
    \vspace{-0.2cm}
\end{figure}

We provide more qualitative results in Fig.~\ref{fig:supp_2}, Fig.~\ref{fig:supp_3}, Fig.~\ref{fig:result-supp}, and Fig.~\ref{fig:supp_5}. 
Note that the images are generated from Stable Diffusion~\cite{rombach2022high}, and the corresponding masks are inferred from our proposed grounding module.
Specifically, the generated images and their corresponding segmentation masks in Fig.~\ref{fig:supp_2} and Fig.~\ref{fig:supp_3}, including common objects, \textit{e.g.}, shark, turtle, and more unusual objects, \textit{e.g.}, Ultraman, pterosaur, Chinese dragon, unicorn and dinosaur, shows the strong generalisability of the grounding module. 
In Fig.~\ref{fig:result-supp}, 
we show more examples from our synthetic semantic segmentation dataset.
In Fig.~\ref{fig:supp_5}, we compare the model trained on our synthesized datasets with other ZS3 methods on PASCAL VOC dataset~\cite{everingham2010pascal}. 
We can observe that the MaskFormer~\cite{cheng2021per} trained on our synthetic semantic segmentation dataset can obtain accurate segmentation on both seen and unseen categories, showing that the guided text-to-image diffusion model can be used to expand the vocabulary of pre-trained detector.

\section{Limitation \& Future Work}
\label{section 6}
In this paper, we have demonstrated the possibility for extracting the visual-language correspondence from a pre-trained text-to-image diffusion model in the form of segmentation map, that can augment the diffusion model with the ability of grounding visual objects along with generation.
However, we also realise there exists certain limitation in this work, 
{\em first}, we only consider to ground the nouns that indicate visual entities,
it would be interesting to ground the human-object, object-object interactions, or even verbs in the future,
{\em second}, we are inserting the grounding module to a pre-trained text-to-image generative model, it would be interesting to co-train the two components, potentially enabling to generate images with higher quality and explainability.

\end{document}